\documentclass{article}

    \PassOptionsToPackage{numbers, compress}{natbib}

    \usepackage[preprint]{neurips_2025}

\usepackage[utf8]{inputenc} %
\usepackage[T1]{fontenc}    %
\usepackage{hyperref}       %
\usepackage{url}            %
\usepackage{booktabs}       %
\usepackage{amsfonts}       %
\usepackage{nicefrac}       %
\usepackage{microtype}      %
\usepackage{xcolor}         %

\usepackage{amsmath}
\usepackage{amssymb}
\usepackage{mathtools}
\usepackage{amsthm}
\theoremstyle{plain}
\newtheorem{theorem}{Theorem}[section]

\theoremstyle{definition}
\newtheorem{definition}[theorem]{Definition}

\theoremstyle{remark}

\usepackage{dsfont}
\usepackage{multirow}
\usepackage{rotating}

\usepackage{algorithm} 
\usepackage{algorithmic}

\usepackage{microtype}
\usepackage{graphicx}
\usepackage{subfigure}
\usepackage{wrapfig}

\usepackage{hyperref}

\title{Action Robust Reinforcement Learning via Optimal Adversary Aware Policy Optimization}

\author{%
  Buqing Nie, Yangqing Fu, Jingtian Ji, Yue Gao \\
  MoE Key Lab of Artificial Intelligence, AI Institute,
  Shanghai Jiao Tong University\\
  \texttt{\{niebuqing, frank79110, jijingtian, yuegao\}@sjtu.edu.cn} \\
  \And
}

\begin{document}

\maketitle

\begin{abstract}
Reinforcement Learning (RL) has achieved remarkable success in sequential decision tasks.
However, recent studies have revealed the vulnerability of RL policies to different perturbations, raising concerns about their effectiveness and safety in real-world applications. 
In this work, we focus on the robustness of RL policies against action perturbations and introduce a novel framework called Optimal Adversary-aware Policy Iteration (OA-PI). 
Our framework enhances action robustness under various perturbations by evaluating and improving policy performance against the corresponding optimal adversaries.
Besides, our approach can be integrated into mainstream DRL algorithms such as Twin Delayed DDPG (TD3) and Proximal Policy Optimization (PPO), improving action robustness effectively while maintaining nominal performance and sample efficiency.
Experimental results across various environments demonstrate that our method enhances robustness of DRL policies against different action adversaries effectively.
\end{abstract}

\section{Introduction}
Recently, Reinforcement Learning (RL)~\cite{sutton2018reinforcement} has achieved remarkable success in sequential decision tasks, including video games~\cite{mnih2015human}, recommendation systems~\cite{afsar2022reinforcement}, and autonomous driving~\cite{zhao2022cadre}.
Besides, Deep Reinforcement Learning (DRL) combines RL with Deep Neural Networks as function approximators, enabling end-to-end learning of policies in continuous control tasks with complex environment dynamics, such as robotic control~\cite{lee2020learning} and conversational systems~\cite{sharma2021towards}.

However, recent works reveal that well-trained Reinforcement Learning policies may be vulnerable to imperceptible perturbations on various aspects, such as observations~\cite{sun2022who}, reward signals~\cite{zhang2020adaptive}, environment dynamics~\cite{rakhsha2020policy}, and multi-agent communications~\cite{sun2022certifiably}.
Previous studies focus primarily on robust RL under observation adversaries,
i.e. the adversary is able to perturb agents' observations under a static constraint~\cite{fischer2019online,zhang2020robust,zhang2021robust}.
Plenty of studies are conducted in this domain, including efficient attack methods~\cite{huang2017adversarial,sun2022who} and the construction of resilient policies as a defense mechanism~\cite{oikarinen2021robust,liang2022efficient,wu2022robust}.

In addition, RL agents are also vulnerable to perturbations in their actions,
resulting in slight deviations from the expected actions prescribed by the policy~\cite{tessler2019action,lee2020spatiotemporally,liu2021provably,pan2022characterizing}.
This phenomenon is prevalent in various application scenarios due to factors such as control errors~\cite{schneider2016improving}, reality gap (modeling discrepancy between simulation and reality)~\cite{zhao2020sim}, external disturbances~\cite{morimoto2005robust}, and adversarial attacks~\cite{lee2020spatiotemporally}.
This susceptibility may affect the effectiveness of the policy and user experience, even causing safety risks in safety-critical tasks such as autonomous driving~\cite{ren2019security} and robot manipulation~\cite{brunke2022safe}.

Several works have been conducted to study the robustness of RL policies against action adversaries. 
Lee et al.~\cite{lee2020spatiotemporally} propose spatiotemporally constrained action space attacks in DRL tasks, which reveals vulnerability of policies to action attacks.
Tessler et al.~\cite{tessler2019action} formulate action robustness as Probabilistic Action Robust MDP (PR-MDP) and Noisy Action Robust MDP (NR-MDP) to model two types of perturbations.
Robust policies are obtained through training alongside action attackers simultaneously.
Liu et al.~\cite{liu2021provably} propose an action poisoning attacker LCB-H, which aims to cheat the agent to learn the policy selected by the attacker in both white-box and black-box settings.

In this paper, we formulate the problem of action robustness in RL as Action-adversarial MDP (AA-MDP), where the action adversaries can add adversarial perturbations to agent's nominal actions directly.
The main contributions of this work are summarized as follows:
\textbf{(1)} 
We propose \emph{Optimal Adversary-aware Policy Iteration (OA-PI)} as a generic framework to enhance policy robustness against action perturbations. 
A novel optimal adversary-aware Bellman operator is introduced, which evaluates the policy performance under the corresponding optimal adversary directly.
\textbf{(2)} The convergence and effectiveness of the OA-PI framework for the AA-MDP problem are theoretically established. 
\textbf{(3)} OA-PI can be incorporated into mainstream DRL algorithms, including TD3~\cite{fujimoto2018addressing} and PPO~\cite{schulman2017proximal}, enhancing action robustness while maintaining nominal performance and sample efficiency.
\textbf{(4)} Experiments on various continuous control tasks are conducted, demonstrating the improvement of our method against various types of action perturbations.

\section{Related Work}

\paragraph{Robust Reinforcement Learning}
Robust reinforcement learning is proposed to enhance performance of RL policies under various perturbations in decision tasks, which is essential to the security of RL in application scenarios~\cite{ilahi2021challenges,pan2022characterizing}.
Plenty of works are conducted to study attack and defense for different types of perturbations,
including dynamics uncertainty~\cite{wang2022policy}, reward perturbations~\cite{eysenbach2022maximum}, domain shift~\cite{muratore2019assessing}, and noisy communication in multi-agent RL~\cite{sun2023certifiably}.
Existing works mainly focus on robustness of policies against state adversaries.
Numerous methods are developed in this domain, including policy regularization~\cite{zhang2020robust,oikarinen2021robust}, training with active attacks~\cite{zhang2021robust}, optimizing worst-case performance~\cite{liang2022efficient}, curriculum learning~\cite{wu2022robust}, and achieving certified robustness~\cite{everett2021certifiable,wu2022crop}.  
In contrast, we focus on action robustness in this work, which is less explored yet critical for real-world applications.

\paragraph{Robust RL under Action Adversaries}
RL policies are also susceptible to action perturbations, especially in environments with complex dynamics, such as robotic control~\cite{lee2020learning,ilahi2021challenges}.
Lee et al.~\cite{lee2020spatiotemporally} propose spatio-temporally constrained action attacks in DRL tasks, demonstrating the vulnerability of policies to action attacks.
Liu et al.~\cite{liu2021provably} propose action poisoning attacks called LCB-H, which aims to cheat the policy to output actions expected by the attacker.
Tessler et al.~\cite{tessler2019action} enhance action robustness of DRL policies based on two novel formulations:
(1) Probabilistic Action Robust MDP, where the agent takes polluted actions with a fixed probability;
(2) Noisy Action Robust MDP, where the agent takes actions interpolated between clean actions and polluted actions.
A new method is proposed to enhance policy robustness through training with adversarial attackers.

\section{Preliminaries}
In this work, we focus on model-free RL with continuous action space.
Typically, the interaction process in RL is formulated as a Markov Decision Process (MDP), denoted as a tuple $\mathcal{M} = <\mathcal{S}, \mathcal{A}, {P}, {R}, \gamma, \rho>$, where $\mathcal{S}$ is the state space, $\mathcal{A}$ is the action space, $P(s'|s,a)$ is the transition probability,
$R:\mathcal{S}\times \mathcal{A}\to \mathbb{R}$ denotes the reward function, $\gamma \in [0,1)$ denotes the discount factor, and $\rho(s)=\Pr(s_0)$ is the distribution of initial states.
The agent takes actions according to its policy, such as  stochastic policy $\pi:\mathcal{S}\to \Pr(a)$ and deterministic policy $\mu:\mathcal{S}\to \mathcal{A}$.
Given a stationary policy $\pi\in\Pi$ ($\Pi$ denotes the policy space), \emph{the nominal performance of $\pi$} can be measured by the value function $V_{\pi}(s)$ and the action value function $Q_{\pi}(s,a)$
:
\begin{equation}
    V_{\pi}(s) \coloneqq \mathbb{E}_{a_t\sim\pi,s_{t+1}\sim P}\left[ \sum_{t=0}^{\infty} \gamma^t r_{t+1} | s_0 = s \right], \; Q_{\pi}(s,a) \coloneqq R(s,a) + \mathbb{E}_{s'\sim P}\left[ V_\pi \left(s'\right) \right].
\label{eq:clean_Value_func}
\end{equation}

\begin{definition}[Bellman Operator]
\label{def:typical_bellman_operator}
For a typical MDP ${\mathcal{M}}$, given a fixed policy $\pi\in\Pi$, the \emph{Bellman Operator} $\mathcal{T}^\pi$ is defined as follows:  
\begin{equation}
\label{eq:typical_bellman_operator}
    \left(\mathcal{T}^\pi Q \right)(s,a) \coloneqq R(s,a) + \gamma \mathbb{E}_{s'\sim P,a'\sim\pi}\left[ Q(s',a') \right].
\end{equation}
\end{definition}
Based on the Bellman operator defined above, the policy evaluation can be conducted to obtain $Q_{\pi}(s,a)$ and $V_{\pi}(s,a)=\mathbb{E}_{a\sim\pi}[Q_{\pi}(s,a)]$.
The agent is trained to find the optimal policy and maximize the cumulative reward, i.e.
$\pi^*\leftarrow\arg\max_{\pi\in\Pi} \, \mathbb{E}_{s\sim \rho} \left[ {V}_{\pi}(s) \right]$.

\section{Methodology}
\label{sec:methodology}
\subsection{Action-adversarial Markov Decision Process}

To study the robustness of RL policies under action perturbations, we formulate the decision process as the action-adversarial Markov Decision Process (AA-MDP), denoted as $\widetilde{\mathcal{M}}$.
Different from typical MDP $\mathcal{M}$, there exists an action adversary $\nu(s)$, which adds perturbations to the agent's actions.

\begin{wrapfigure}{R}{0.55\textwidth}
    \centering
    \vskip -0.10cm
    \includegraphics[width=0.55\textwidth]{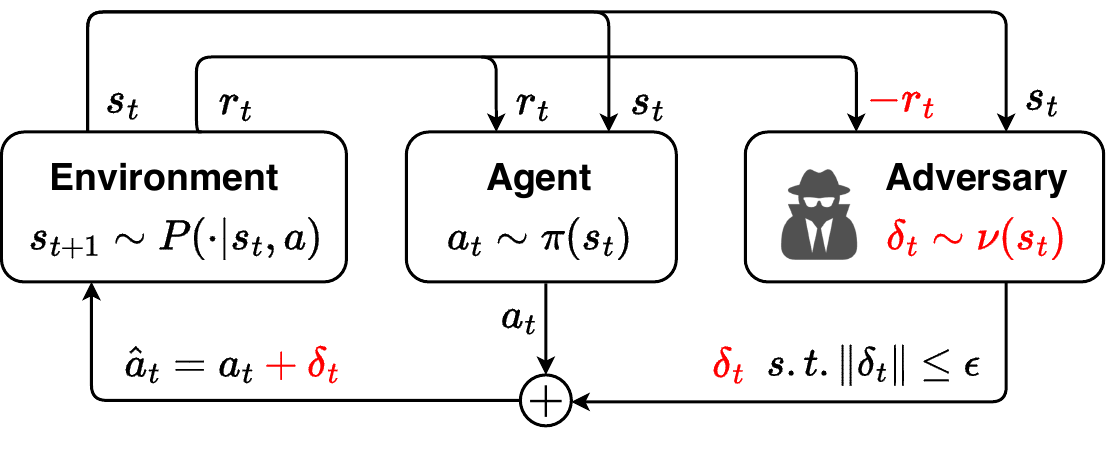}
    \vskip -0.05cm
    \caption{The interaction process between the agent and environment in AA-MDP. 
    The agent takes actions $\hat{a}_t=a_t + \delta_t$ polluted by the adversary $\nu(s)$ rather than clean actions $a_t$.
    The perturbation $\delta_t$ is bounded by $\epsilon$.}
    \label{fig:interaction_process}
    \vskip -0.45cm
\end{wrapfigure}
As shown in Fig.~\ref{fig:interaction_process}, each time the agent obtains the observation $s_t$ and makes the decision $a_t\sim\pi(s_t)$.
The action adversary $\nu$ generates action perturbations $\delta_t\sim\nu(s_t)$, resulting in the agent taking the perturbed action $\hat{a}_t=a_t + \delta_t$, which may be unreasonable and cause safety risks.
In this work, the action perturbations $\delta_t$ are $l_\infty$ bounded, i.e. $\|\delta_t\|_{\infty} \leq \epsilon$, where $\epsilon\geq 0$ is an important parameter determining the maximum strength of the action perturbations. 
Larger values of $\epsilon$ indicate stronger adversaries, which in turn require a higher level of the policy robustness.

Similar to Eq.~\eqref{eq:clean_Value_func} in typical MDP $\mathcal{M}$, we can formulate the value function and action-value function in AA-MDP $\widetilde{\mathcal{M}}$ for fixed policy $\pi$ and adversary $\nu$:
\begin{equation}
    V_{\pi\circ\nu}(s) \coloneqq \mathbb{E}_{\pi,\nu,P}\left[ \sum_{t=0}^{\infty} \gamma^t r_{t+1} | s_0 = s \right], \;
    Q_{\pi\circ\nu}(s,a) \coloneqq  R(s,a) + \gamma  \mathbb{E}_{P}\left[ V_{\pi\circ\nu}(s')
    \right].
\label{eq:adversary_V_func}
\end{equation}

Given a fixed policy $\pi$ in AA-MDP $\widetilde{\mathcal{M}}$, the objective of adversary $\nu$ is to minimize the expected discounted return of the agent.
The goal of AA-MDP is to improve the robustness of the policy through maximizing discounted return against the corresponding optimal adversary $\nu^*$.
Therefore, our task can be formulated as the following optimization problem:
\begin{equation}
\label{eq:optim_problem}
\max_{\pi\in\Pi} \, \min_{\nu} \, \mathbb{E}_{s\sim \rho} \left[ V_{\pi\circ\nu}(s) \right] \;\, \text {s.t.} \;\,
\|\delta\|_{\infty} \leq \epsilon, \, \forall s\in \mathcal{S}, \, \delta \sim\nu(s).
\end{equation}

\subsection{Optimal Adversary-Aware Policy Iteration Algorithm}
As described in Eq.~\eqref{eq:optim_problem}, we are required to solve a maximin optimization problem, which 
involves finding optimal action adversaries $\nu^*=\arg\min_\nu \mathbb{E}_{s\sim \rho} \left[ V_{\pi\circ\nu}(s) \right]$ given fixed policies $\pi$.
Finding $\nu^*$ is equivalent to learning the optimal policy in another MDP, where the adversary $\nu$ is trained to minimize the episode reward of the agent, considering $\pi$ as part of the environment.
More details are given in Appendix~\ref{sec:app_MDP_adversary}.
However, learning such an optimal adversary $\nu^*$ requires additional sampling and computational cost, which is quite expensive or unavailable in some applications.
In this section, we propose \emph{optimal adversary (OA)-aware Policy Iteration} without formulating or learning the adversary explicitly to address this issue. 

\subsubsection{OA-Aware Policy Evaluation}
Typical RL utilizes $V_{\pi}(s)$ and $Q_{\pi}(s,a)$ described in Eq.~\eqref{eq:clean_Value_func} to evaluate the policy $\pi$ in typical MDP $\mathcal{M}$, without considering the influence of adversary $\nu$.
In order to evaluate the performance of policy $\pi$ against its optimal action adversary $\nu^*$, we construct the \emph{optimal adversary-aware Bellman Operator} as follows to find $Q_{\pi\circ\nu^*}$ and $V_{\pi\circ\nu^*}$.

\begin{definition}%
\label{def:bellman_operator}
For an AA-MDP $\widetilde{\mathcal{M}}$, given a fixed policy $\pi\in\Pi$ and the strength $\epsilon\geq 0$ of the adversary $\nu$, the \emph{optimal adversary-aware Bellman operator} $\widetilde{\mathcal{T}}^\pi$ is defined as follows:  
\begin{equation}
\label{eq:bellman_operator}
    \big(\widetilde{\mathcal{T}}^\pi Q_{\operatorname{adv}}\big)(s,a) \coloneqq R(s,a) + \gamma \mathbb{E}_{s'\sim P} \left[\min_{\|\delta\|\leq \epsilon} \mathbb{E}_{a'\sim\pi(s')}\left[ Q_{\operatorname{adv}}(s',a'+\delta) \right]\right].
\end{equation}
\end{definition}

Intuitively, 
the minimization operator, as illustrated in Eq.~\eqref{eq:bellman_operator}, is employed to calculate the worst performance over future steps while considering action perturbations constrained by $\epsilon$. 
Note that the perturbations $\delta$ are strategically planned and temporally correlated to systematically minimize the performance of $\pi$ throughout the entire episode.

\begin{theorem}
\label{thm:bellman_operator_convergence}
Given any fixed policy $\pi \in\Pi$ with $\epsilon\geq 0$, the optimal adversary-aware Bellman operator $\widetilde{\mathcal{T}}^\pi$ is a contraction mapping under sup-norm $\|\cdot\|_{\infty}$, with the fixed point $Q_{\pi\circ\nu^*}$, i.e. $\lim_{k\to\infty}(\widetilde{\mathcal{T}}^\pi)^k Q_{\operatorname{adv}} = Q_{\pi\circ\nu^*}$.
\end{theorem} 

Theorem~\ref{thm:bellman_operator_convergence} reveals that the $Q_{\operatorname{adv}}$ described in Eq.~\eqref{eq:bellman_operator} approaches to $Q_{\pi\circ\nu^*}$ after sufficient iterations, thus can be utilized to evaluate the performance of policy $\pi$ under the corresponding optimal adversary $\nu^*$.
The detailed proof of  this theorem is given in Appendix~\ref{sec:app_proof_bellman_operator_convergence}. 
Thus, given $\forall \pi\in\Pi, \epsilon\geq 0$, we can evaluate the value of $\pi$ under the optimal adversary:
\begin{equation}
\label{eq:adversary_Q_to_V_func}
V_{\pi\circ\nu^*}(s) =
\min_{\|\delta\|\leq\epsilon} \mathbb{E}_{\pi}\left[ Q_{\pi\circ\nu^*}\left(s,a+\delta \right) \right]
= \lim_{k\to\infty}\min_{\|\delta\|\leq\epsilon} \mathbb{E}_{\pi}\left[ \left(\widetilde{\mathcal{T}}^\pi\right)^k Q_{\operatorname{adv}}\left(s, a+\delta\right) \right],
\end{equation}
which is the optimization objective of policy $\pi$ described in the optimization problem Eq.~\eqref{eq:optim_problem}.

\subsubsection{OA-Aware Policy Improvement}
As the optimization problem illustrated in Eq.~\eqref{eq:optim_problem}, the policy is trained to optimize the performance under the optimal action adversary $V_{\pi\circ\nu^*}$.
As described in Eq.~\eqref{eq:bellman_operator} and Eq.~\eqref{eq:adversary_Q_to_V_func}, $V_{\pi\circ\nu^*}$ can be obtained utilizing OA-aware Bellman operator.
Thus, we can conduct policy improvement as follows:
\begin{equation}
\label{eq:policy_improvement}
    \pi \leftarrow \arg\max_{\pi\in\Pi}\mathbb{E}_{s\sim\rho}\left[V_{\pi\circ\nu^*}(s)\right]
    = \arg\max_{\pi\in\Pi} \min_{\|\delta\|\leq\epsilon} \mathbb{E}_{s\sim\rho,a\sim\pi}\left[Q_{\operatorname{adv}}\left(s,a+\delta\right)\right].
\end{equation}

\begin{theorem}[Policy Improvement Theorem in AA-MDP]
\label{thm:policy_improvement_thm}
Given any two policies $\pi$ and $\pi'$, 
we define $Q_{\operatorname{adv}}^{\pi} (s,\pi') = \min_{\| \delta \| \leq \epsilon} \mathbb{E}_{a\sim\pi'}[ Q_{\operatorname{adv}}^{\pi}(s,a+\delta) ]$.
For $\forall s\in\mathcal{S}$, if $Q_{\operatorname{adv}}^{\pi} (s,\pi')\geq V_{\pi\circ\nu^*}(s)$, then we have $V_{\pi'\circ\nu^*}(s) \geq V_{\pi\circ\nu^*}(s)$ for $\forall s\in\mathcal{S}$, i.e. $\pi'$ is at least good as policy $\pi$.
\end{theorem}

The detailed proof is given in Appendix~\ref{sec:app_policy_improvement_thm}.
Theorem~\ref{thm:policy_improvement_thm} can be utilized to prove the effectiveness of the policy update described in Eq.~\eqref{eq:policy_improvement}, which is illustrated as follows.

\begin{theorem}
\label{thm:policy_improvement}
Given $\forall \epsilon\geq 0$, $\pi$ and $\pi'$ denote policies before and after optimal adversary-aware policy improvement, we have $V_{\pi'\circ\nu^*}(s)\geq V_{\pi\circ\nu^*}(s)$ for $\forall s\in \mathcal{S}$.
\end{theorem} 

Theorem~\ref{thm:policy_improvement} demonstrates that the policy achieves a monotonically nondecreasing update through the policy improvement described in Eq.~\eqref{eq:policy_improvement}.
In other words, the updated policy is guaranteed to be at least as good as the previous policy.
The detailed proof is given in Appendix~\ref{sec:app_proof_policy_improvement}. 

\subsubsection{OA-Aware Policy Iteration}

In this section, we propose \emph{optimal adversary-aware Policy Iteration} (OA-PI).
As depicted in Algorithm~\ref{alg:policy_iteration_algo}, OA-PI alternatively performs OA-aware policy evaluation and policy improvement until convergence, in accordance with Eq.~\eqref{eq:bellman_operator} and Eq.~\eqref{eq:policy_improvement}.

\begin{theorem}
\label{thm:policy_iteration_convergence}
Given $\forall Q_{\operatorname{adv}}, \pi\in\Pi$ for initialization, the policy will converge to the optimal policy $\pi^*$ after sufficient optimal adversary-aware Policy Iterations.
\end{theorem} 
\begin{wrapfigure}{R}{0.65\textwidth}
    \centering
    \vskip -0.38cm
    \begin{minipage}{0.65\textwidth}
        \begin{algorithm}[H]
        \caption{Optimal adversary-aware Policy Iteration}
        \label{alg:policy_iteration_algo}
        \begin{algorithmic}[1]
           \STATE {\bfseries Input:} adversary strength $\epsilon$.
           \STATE {\bfseries Output:} robust policy $\pi$.
           \STATE Initialize policy $\pi$ and $Q_{\operatorname{adv}}$ randomly.
           \REPEAT 
                \STATE $\displaystyle Q_{\operatorname{adv}}(s,a) \leftarrow R(s,a) + \gamma \mathbb{E}_{P} \min_{\|\delta\|\leq \epsilon} \mathbb{E}_{\pi}\left[ Q_{\operatorname{adv}}(s',a'+\delta) \right]$
                \STATE $\displaystyle \pi\leftarrow\arg\max_{\pi} \min_{\|\delta\|\leq\epsilon} \mathbb{E}_{\rho,\pi}\left[Q_{\operatorname{adv}}\left(s,a+\delta\right)\right]$
           \UNTIL{convergence.}
        \end{algorithmic}
        \end{algorithm}
    \end{minipage}
    \vskip -0.3cm
\end{wrapfigure}

This theorem is obtained based on Theorem~\ref{thm:bellman_operator_convergence} and \ref{thm:policy_improvement}.
The detailed proof is given in Appendix~\ref{sec:app_proof_policy_iteration_convergence}.
Theorem~\ref{thm:policy_iteration_convergence} indicates that we can solve the optimization problem shown in Eq.~\eqref{eq:optim_problem} through sufficient iterations.
OA-PI is a generic framework to enhance policy robustness against action perturbations in AA-MDP, thus can be applied to mainstream DRL methods.

\subsection{Optimal Adversary-Aware DRL}
\label{sec:method_OA_DRL}
Based on the OA-PI algorithm presented in the previous section, we further derive practical implementations on DRL for continuous control tasks.
OA-PI is a generic training framework for robust reinforcement learning against action adversaries, thus can be combined with mainstream DRL algorithms, including on-policy/off-policy DRL with stochastic/deterministic policies.

\subsubsection{Combined with Stochastic Policies}
\label{sec:oa_ppo}
In this section, we 
propose \emph{optimal adversary-aware PPO (OA-PPO)} through incorporating OA-PI concept into Proximal Policy Optimization (PPO) algorithm~\cite{schulman2017proximal}.

\textbf{Policy evaluation:}
In addition to typical critics utilized in PPO, there exists a new OA-aware critic $Q_{\operatorname{adv}}(s,a)$ in OA-PPO algorithm, which evaluates the robustness of $\pi$ against action adversaries.
$Q_{\operatorname{adv}}(s,a)$ is trained utilizing Mean Square Error (MSE) loss with labels obtained through the OA-aware Bellman operator depicted in Eq.~\eqref{eq:bellman_operator}.

\textbf{Policy improvement:}
The policy $\pi$ is updated through the following formulation:
\begin{equation}
\label{eq:stochastic_policy_improvement} 
{\pi}\leftarrow \arg
\max_{\pi} 
\mathbb{E}_{(s,a)\sim\mathcal{D}}\left[ 
\min\left( 
\rho_{\pi} \widetilde{A}(s,a), \,
g(\rho_{\pi}) \widetilde{A}(s,a) 
\right)
\right],
\end{equation}
where $\mathcal{D}$ denotes the replay buffer, $\rho_{\pi}=\frac{\pi(a|s)}{\pi_{\operatorname{old}}(a|s)}$, and $g(\rho_{\pi}) = \operatorname{clip}(\rho_{\pi}, 1-\xi, 1+\xi)$. $\xi$ is a small hyper-parameter to limit the magnitude of the update, promoting stable and controlled updates.
$\widetilde{A} (s,a) = \omega A(s,a) + (1-\omega) Q_{\operatorname{adv}}(s,a+\delta^*)$, where $A(s,a)$ denotes the advantage estimation of the current policy $\pi_{\operatorname{old}}$, and $\delta^* = \arg\min_{\|\delta\|\leq \epsilon} Q_{\operatorname{adv}}(s,a+\delta)$ denotes the optimal action perturbations, which can be obtained through gradient decent-based methods approximately, such as Projected Gradient Descent (PGD)~\cite{madry2018towards} used in this work.
The hyper-parameter $\omega\in [0,1]$ is utilized to take a trade-off between optimality (nominal performance) and action robustness (performance under perturbations) of the policy.

Based on the above descriptions, we can construct the OA-PPO algorithm.
More details including the pseudo-code (Algorithm~\ref{alg:oa_ppo}) are shown in the experiment section and Appendix~\ref{sec:app_oa_ppo_algo}.

\subsubsection{Combined with Deterministic Policies}
\label{sec:oa_td3}

\begin{algorithm}[tb]
   \caption{Optimal adversary-aware TD3}
   \label{alg:oa_td3}
\begin{algorithmic}[1]
   \STATE {\bfseries Input:} adversary strength $\epsilon$, weight $\omega$.
   \STATE {\bfseries Initialize:} actor $\mu$, $\mu'$, buffer $\mathcal{D}$, typical critics $Q_{\{1,2\}}$, $Q_{\{1,2\}}'$, OA-aware critic $Q_{\operatorname{adv}}$, $Q_{\operatorname{adv}}'$.
   \FOR{$t=1$ \textbf{to} $T$}
    \STATE Execute $a_t = \mu(s_t) + \xi$ and observe $s_{t+1}$ with reward $r_t$, where $\xi\sim\mathcal{N}$.
    \STATE Store transition $(s_t, a_t, r_t, s_{t+1},d_t)$ into buffer $\mathcal{D}$
    \STATE Sample a batch of data $B=\left\{ \left( s,a,r,s',d \right) \right\}$ from $\mathcal{D}$
    \STATE Compute target action $a'= \mu'(s')+\xi$, where $\xi\sim\operatorname{clip}\left(\mathcal{N},-c,c \right)$
    \STATE $\theta^{Q_{i}}\leftarrow \arg\min \mathbb{E}_{ B} (y - Q_{i}(s,a))^2$, \, $y=r+\gamma (1-d)\min_{i}Q_{i}'(s',a')$
    \STATE $\theta^{Q_{\operatorname{adv}}}\leftarrow \arg\min \mathbb{E}_{ B} (y_{\operatorname{adv}} - Q_{\operatorname{adv}}(s,a))^2$, \, $ y_{\operatorname{adv}} = r+\gamma (1-d)\min_{\|\delta\|\leq \epsilon} Q_{\operatorname{adv}}'(s',a'+\delta)$
    \IF{ $t\mod\operatorname{policy\_delay}$}
        \IF{\textcolor{black}{$(\nabla_{\theta^{\mu}}Q_1)\cdot (\nabla_{\theta^{\mu}}Q_{\operatorname{adv}}) < 0$}}
            \STATE \textcolor{black}{Update $\theta^\mu$ using $\omega\operatorname{proj}\left( \nabla_{\theta^{\mu}}Q_1 \to \nabla_{\theta^{\mu}}Q_{\operatorname{adv}} \right) + (1-\omega)\operatorname{proj}\left( \nabla_{\theta^{\mu}}Q_{\operatorname{adv}} \to \nabla_{\theta^{\mu}}Q_1 \right)$}
        \ELSE
            \STATE \textcolor{black}{Update $\theta^\mu$ using $\omega \nabla_{\theta^{\mu}}Q_1 + (1-\omega) \nabla_{\theta^{\mu}}Q_{\operatorname{adv}}$}
        \ENDIF
        \STATE Update target networks  $\mu'$, $Q_{i\in\{1,2\}}'$, and $Q_{\operatorname{adv}}'$ based on soft updates.
    \ENDIF
    
   \ENDFOR
\end{algorithmic}
\end{algorithm}

As illustrated in Algorithm~\ref{alg:oa_td3}, we propose \emph{optimal adversary-aware TD3 (OA-TD3)} algorithm based on Twin Delayed DDPG (TD3)~\cite{fujimoto2018addressing} in this section.
To ensure clarity and generality, we describe optimal adversary-aware concepts within general deterministic policy gradient methods, rather than focusing solely on TD3.
$\mu$ is used to represent deterministic policies instead of $\pi$ for distinction.

\textbf{Policy evaluation:}
Two critics $Q(s,a)$ and $Q_{\operatorname{adv}}(s,a)$ are utilized to evaluate the nominal performance and policy robustness of $\mu$ under action adversaries.
Both of two critics are trained based on MSE loss.
Labels for $Q(s,a)$ and $Q_{\operatorname{adv}}(s,a)$ are obtained according to the typical Bellman operator and OA-aware Bellman operator depicted in Eq.~\eqref{eq:typical_bellman_operator} and Eq.~\eqref{eq:bellman_operator} respectively.

\textbf{Policy improvement:} 
The policy $\mu$ is trained based on deterministic policy gradient with two critics $Q(s,a)$ and $Q_{\operatorname{adv}}(s,a)$ simultaneously:
\begin{equation}
\label{eq:deterministic_policy_improvement}
\mu \leftarrow \max_{\mu}\mathbb{E}_{s\sim\mathcal{D}}\left[\omega Q\left(s,\mu\left(s\right)\right) + (1-\omega) Q_{\operatorname{adv}}\left(s,\mu\left(s\right)+\delta^*\right) \right],
\end{equation}
where $\delta^* = \arg\min_{\|\delta\|\leq \epsilon} Q_{\operatorname{adv}}(s,\mu(s)+\delta)$ is the optimal action perturbation.
$Q$ and $Q_{\operatorname{adv}}$ represent two optimization objectives: optimality (nominal performance) and action robustness accordingly.
$\omega$ is a hyper-parameter and takes a trade-off between two objectives derived from two critics. 

However, there exist conflicts between policy gradients of above two objectives.
Formally, gradients of $\theta^{\mu}$ derived from two critics are defined to be conflicting if they have negative cosine similarity, i.e. $(\nabla_{\theta^{\mu}}Q)\cdot (\nabla_{\theta^{\mu}}Q_{\operatorname{adv}}) < 0$.
This conflict could impede the optimization process, resulting in training inefficiency and a significant reduction in policy performance.

To address this issue, we employ Gradient Surgery (GS)~\cite{yu2020gradient} to resolve conflicts by gradient projection.
As shown in Algorithm~\ref{alg:oa_td3}, when conflicts between $\nabla_{\theta^{\mu}}Q$ and $\nabla_{\theta^{\mu}}Q_{\operatorname{adv}}$ are observed, we resolve conflicts by projecting each gradient onto the normal plane of the other.
Thus, $\mu$ is trained using projected gradients, i.e. $\operatorname{proj}\left( \nabla_{\theta^{\mu}}Q \to \nabla_{\theta^{\mu}}Q_{\operatorname{adv}} \right)$ and $\operatorname{proj}\left( \nabla_{\theta^{\mu}}Q_{\operatorname{adv}} \to \nabla_{\theta^{\mu}}Q \right)$, where 
$\operatorname{proj}\left(\mathbf{g}_i\to\mathbf{g}_j\right) = \mathbf{g}_i - \frac{\mathbf{g}_i\cdot \mathbf{g}_j}{\| \mathbf{g}_j\|^2}\mathbf{g}_j$.
More details are given in Appendix~\ref{sec:app_pcgrad}.

\section{Experiments}
\label{sec:experiments}
In this section,  we conduct various experiments to study the following questions:
\textbf{(a)} Can OA-DRL enhance robustness against different perturbations?
\textbf{(b)} Can OA-DRL maintain nominal performance and sample efficiency compared to typical DRL methods?
\textbf{(c)} Is OA-aware concept effective for different DRL, including on/off-policy methods with stochastic/deterministic policies?

In this work, OA-aware concept is applied to PPO and TD3 algorithms on the following tasks:
\textbf{(a) Box2d tasks:} Two classic Box2d-based control tasks, including \emph{Bipedal Walker} and \emph{Lunar Lander}.
\textbf{(b) MuJoCo tasks:} Five locomotion tasks based on the MuJoCo engine~\cite{todorov2012mujoco} with complex environment dynamics, including \emph{Walker2d}, \emph{Hopper}, \emph{Humanoid}, \emph{Ant}, and \emph{HalfCheetah}.

\subsection{Experiments for Stochastic Policies}

\subsubsection{Experiment Settings}
\label{sec:exp_settings_stochastic}

\textbf{Baselines:}
In this work, we compare the OA-PPO algorithm with the following baseline methods, including
\textbf{(a) PPO}: typical PPO algorithm; 
\textbf{(b) PPO-Noise}: training PPO in polluted environments with random action noises at each step, which is commonly utilized to solve Sim2Real gaps, such as domain randomization for robotics control~\cite{rudin2022learning}. 
\textbf{(c) PPO-Min-Q}: replace OA-aware critic $Q_{\operatorname{adv}}$ in OA-PPO with typical Q functions.

\textbf{Metrics:} 
In this experiment, the following adversaries are utilized to conduct action attacks on the policies.
The average episode return is recorded to evaluate the policy performance.
\textbf{(a) nominal attacks:} the agent executes clean actions at each step, i.e. $\delta=0$, which is utilized to evaluate the policy's nominal performance.
\textbf{(b) Random attacks}: adding random perturbations $\delta\sim U[-\epsilon, \epsilon]$ to the actions taken by the agent at each step.
\textbf{(c) Biggest attacks}: adding the biggest perturbations $\delta\in\{ -\epsilon, \epsilon\}$ randomly at each step.
\textbf{(d) Min-Q attacks}: adding perturbations with minimized Q values, i.e. $\delta= \arg\min_{\delta} Q(s, a+\delta)$.
\textbf{(e) Min-OA-Q attacks}: adding perturbations with minimized OA-aware Q values, i.e. $\delta= \arg\min_{\delta} Q_{\operatorname{adv}}(s, a+\delta)$.
To perform Min-Q and Min-OA-Q  attacks fairly, dedicated $Q$ and $Q_{\operatorname{adv}}$ functions are trained for each policy network following Definition~\ref{def:typical_bellman_operator} and \ref{def:bellman_operator}.
More details of the Min-OA-Q attack are given in Appendix~\ref{sec:app_min_oa_q_attacks}.

In this experiment, we implement OA-PPO based on the PPO implementation proposed in \cite{huang202237}.
Each method is trained 2M steps in most environments with at least 10 random seeds.
The $\epsilon$ value denotes the perturbation strength during robust training, which is tuned in $[0.1,0.3]$ according to different tasks.
The hyper-parameter $\omega$ is utilized to take a balance between optimality (nominal performance) and robustness (performance against perturbations), which is tuned in $[0.4, 0.6]$ given different tasks.
The optimal one-step perturbations $\delta^*$ utilized in policy evaluation Eq.~\eqref{eq:bellman_operator} and policy improvement Eq.~\eqref{eq:stochastic_policy_improvement} are obtained though multi-step $l_\infty$ $\epsilon$-bounded PGD attacks.

\begin{table}[t]
\caption{Experiment results on robustness of stochastic polices under different action perturbations,
including the nominal performance without attacks.
The episode returns with standard errors are obtained over 10 seeds and 100 episodes.
The best results are boldfaced.
}
\label{tab:rob_res_ppo}
\centering
\begin{tabular}{llccccc}
\toprule
Task & Method & Nominal & Random & Biggest & Min-Q & Min-OA-Q \\ 
\midrule
\multirow{4}{*}{Ant} 
& PPO & 2964$\pm$232 & 2429$\pm$242 & 1876$\pm$250 & 540$\pm$205 & -114$\pm$217 \\ 
& PPO-Noise & 2760$\pm$361 & 2387$\pm$379 & 1753$\pm$288 & 764$\pm$219 & 58$\pm$192 \\ 
& PPO-Min-Q & 2803$\pm$228 & 2381$\pm$357 & 1620$\pm$252 & 578$\pm$204 & -40$\pm$178 \\ 
& \textbf{OA-PPO}& \textbf{3135$\pm$245} & \textbf{2588$\pm$225} & \textbf{1961$\pm$139} & \textbf{890$\pm$241} & \textbf{523$\pm$210} \\ 
\midrule
\multirow{4}{*}{Walker2d} 
& PPO & 4324$\pm$351 & 4283$\pm$320 & 3992$\pm$267 & 2511$\pm$370 & 1319$\pm$425 \\ 
& PPO-Noise & 4366$\pm$202 & 4393$\pm$171 & 4179$\pm$270 & 2621$\pm$446 & 899$\pm$73 \\ 
& PPO-Min-Q & 3993$\pm$336 & 3822$\pm$298 & 3596$\pm$493 & 2501$\pm$357 & 1415$\pm$356 \\ 
& \textbf{OA-PPO}& \textbf{4705$\pm$283} & \textbf{4650$\pm$172} & \textbf{4538$\pm$144} & \textbf{3291$\pm$393} & \textbf{1936$\pm$381} \\ 
\midrule
\multirow{4}{*}{Hopper} 
& PPO & 3180$\pm$214 & 2916$\pm$329 & 2713$\pm$223 & 1551$\pm$388 & 1473$\pm$170 \\ 
& PPO-Noise & 3108$\pm$357 & 2951$\pm$324 & 2668$\pm$336 & 2092$\pm$349 & 1773$\pm$192 \\ 
& PPO-Min-Q & 3240$\pm$173 & 3044$\pm$238 & 2921$\pm$159 & 2397$\pm$302 & 1852$\pm$169 \\ 
& \textbf{OA-PPO}& \textbf{3413$\pm$208} & \textbf{3379$\pm$159} & \textbf{3244$\pm$212} & \textbf{2967$\pm$251} & \textbf{2575$\pm$144} \\ 
\bottomrule
\end{tabular}
\end{table}

\subsubsection{Experiment Results}

The experiment results on action robustness are shown in Table~\ref{tab:rob_res_ppo}.
As illustrated in the table, OA-PPO outperforms baseline methods across both nominal cases and different action adversaries in various environments, demonstrating that OA-aware policy improvement enhances policy robustness effectively.
Take \emph{Walker2d} task as an example, OA-PPO achieves an episode return of 1936 under the Min-OA-Q adversary, outperforming the best baseline method's score 1415 by approximately $36.8\%$.
More experiment results are shown in Appendix~\ref{sec:app_addtional_res_ppo}. 

As shown in the table, Min-OA-Q achieves stronger attacks than other adversaries, leading to the worst performance of policies under different adversaries.
The Min-OA-Q attack is conducted based on OA-aware Bellman operator, which learns the optimal adversary implicitly via evaluating the worst performance under perturbations.

\begin{wrapfigure}{R}{0.39\textwidth}
    \centering
    \vskip -0.3cm
    \includegraphics[width=0.39\textwidth]{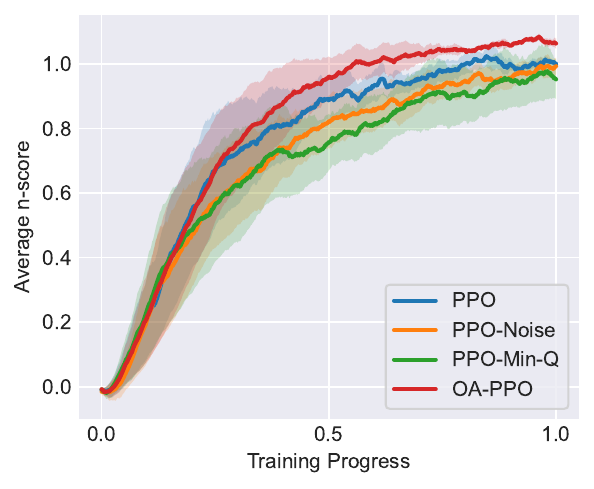}
    \vskip -0.1cm
    \caption{Normalized training curves of OA-PPO and baseline methods.
    See Fig.~\ref{fig:learning_curves_ppo} in Appendix~\ref{sec:app_addtional_res_ppo} for individual training curves in each task. }
    \label{fig:learning_curve_ppo_all_norm}
    \vskip -0.5cm
\end{wrapfigure}

In addition, the training curves of each method are shown in Fig.~\ref{fig:learning_curve_ppo_all_norm}, with $x$-axis representing training progress and $y$-axis indicating \emph{normalized nominal performance (n-score)} computed across diverse tasks.
Learning curves in each environment are shown in Appendix~\ref{sec:app_addtional_res_ppo}.
The detailed computation process of \emph{n-score} is described in Appendix~\ref{sec:app_normalize_reward}.
The lines denote the mean score, while shaded regions represent standard errors across multiple random seeds.

As illustrated in the figure, OA-PPO (red lines) exhibits comparable or even superior sample efficiency compared to other methods including typical PPO algorithm.
This intriguing phenomenon suggests that action robustness and high sample efficiency can be achieved concurrently within DRL domains, rather than being inherently contradictory.

As illustrated in Table~\ref{tab:rob_res_ppo} and Fig.~\ref{fig:learning_curve_ppo_all_norm}, PPO-Min-Q achieves worse performance than OA-PPO generally in each environment with different adversaries,  demonstrating the effectiveness and necessity of OA-aware critic in our method. 
Besides, the improvement of PPO-Noise is limited and unstable across different tasks and adversaries.
In tasks such as \emph{Ant}, PPO-Noise enhances robustness under Min-Q and Min-OA-Q adversaries, at the expense of reduced sample efficiency and inferior performance under other perturbations  compared to PPO.
This suggests that while active environmental pollution during training can improve policy robustness, the approach is unstable and may degrade sample efficiency and overall policy performance.

\subsection{Experiments for Deterministic Policies}
\label{sec:experiment_oa_td3}

\begin{table}[tbp]
\caption{Experiment results on robustness of deterministic polices under different action perturbations.
}
\label{tab:rob_res_td3}
\begin{center}
\begin{tabular}{llccccc}
\toprule
Task & Method & Nominal & Random & Biggest & Min-Q & Min-OA-Q \\ \midrule
\multirow{4}{*}{Ant} 
& TD3 & 5548$\pm$815 & 5447$\pm$215 & 4656$\pm$344 & 3435$\pm$418 & 719$\pm$529 \\  
& TD3-Noise & 5330$\pm$276 & 4766$\pm$364 & 4360$\pm$289 & 3283$\pm$429 & 2130$\pm$456 \\ 
& NR/PR-MDP & 705$\pm$165 & 439$\pm$113 & 411$\pm$147 & 184$\pm$106 & 56$\pm$179 \\ 
& \textbf{OA-TD3} & \textbf{6228$\pm$106} & \textbf{5793$\pm$317} & \textbf{5201$\pm$407} & \textbf{4928$\pm$360} & \textbf{2853$\pm$383} \\ 
& Ours w/o GS & 5587$\pm$299 & 5258$\pm$412 & 4925$\pm$385 & 4565$\pm$281 & 2127$\pm$455 \\ 
\midrule
\multirow{4}{*}{Wal.} 
& TD3 & 3953$\pm$358 & 3925$\pm$321 & 3581$\pm$238 & 1938$\pm$438 & 740$\pm$281 \\  
& TD3-Noise & 3830$\pm$417 & 3790$\pm$385 & 3819$\pm$368 & 3189$\pm$370 & 1064$\pm$384 \\ 
& NR/PR-MDP & 2616$\pm$533 & 2571$\pm$344 & 2467$\pm$430 & 2389$\pm$313 & 1070$\pm$229 \\ 
& \textbf{OA-TD3}& \textbf{4725$\pm$151} & \textbf{4713$\pm$190} & \textbf{4637$\pm$248} & \textbf{4006$\pm$383} & \textbf{1172$\pm$317} \\  
& Ours w/o GS & 4227$\pm$480 & 4224$\pm$503 & 4184$\pm$483 & 3940$\pm$582 & 1068$\pm$392 \\ 
\midrule
\multirow{4}{*}{Hop.} 
& TD3 & 3155$\pm$437 & 2145$\pm$561 & 2134$\pm$537 & 1110$\pm$176 & 621$\pm$50 \\  
& TD3-Noise & 2887$\pm$630 & 2814$\pm$564 & 2519$\pm$495 & 1149$\pm$380 & 778$\pm$152 \\ 
& NR/PR-MDP & 2865$\pm$92 & 2830$\pm$142 & 2791$\pm$353 & 2381$\pm$147 & 806$\pm$170 \\ 
& \textbf{OA-TD3}& \textbf{3618$\pm$78} & \textbf{3433$\pm$90} & \textbf{3174$\pm$261} & \textbf{2554$\pm$342} & \textbf{914$\pm$147} \\  
& Ours w/o GS & 3073$\pm$614 & 3050$\pm$412 & 2467$\pm$580 & 1025$\pm$257 & 812$\pm$75 \\ 
\midrule
\multirow{4}{*}{HC.} 
& TD3 & 9745$\pm$663 & 8985$\pm$678 & 7661$\pm$649 & 5984$\pm$221 & 2941$\pm$416 \\  
& TD3-Noise & 11167$\pm$204 & 10485$\pm$271 & 8802$\pm$174 & 6379$\pm$326 & 2902$\pm$514 \\ 
& NR/PR-MDP & 1461$\pm$231 & 1390$\pm$220 & 1357$\pm$112 & 1323$\pm$402 & 676$\pm$348 \\ 
& \textbf{OA-TD3}& \textbf{12088$\pm$161} & \textbf{11036$\pm$123} & \textbf{9519$\pm$200} & \textbf{7786$\pm$437} & \textbf{3228$\pm$424} \\  
&  Ours w/o GS & 10458$\pm$277 & 9575$\pm$330 & 8181$\pm$481 & 6436$\pm$429 & 2954$\pm$487 \\ 
\bottomrule
\end{tabular}
\end{center}
\end{table}

\subsubsection{Experiment Settings}
\label{sec:exp_settings_deterministic}

\textbf{Baselines:}
In this work, we compare the OA-TD3 algorithm with the following methods, including
\textbf{(a) TD3}: typical TD3 algorithm; 
\textbf{(b) TD3-Noise}: training TD3 in polluted environments with random action noises at each step.
\textbf{(c) NR/PR-MDP}~\cite{tessler2019action}: training robust policies alongside adversarial action attackers in NR-MDP and PR-MDP.
In this experiment, the higher scores of two baseline methods are reported for each task and adversary.
\textbf{(d) OA-TD3 w/o GS}: training OA-TD3 without Gradient Surgery, which aims to demonstrate the effectiveness of GS illustrated in Section~\ref{sec:oa_td3}.

\begin{wrapfigure}{R}{0.39\textwidth}
    \centering
    \vskip -0.4cm
    \includegraphics[width=0.39\textwidth]{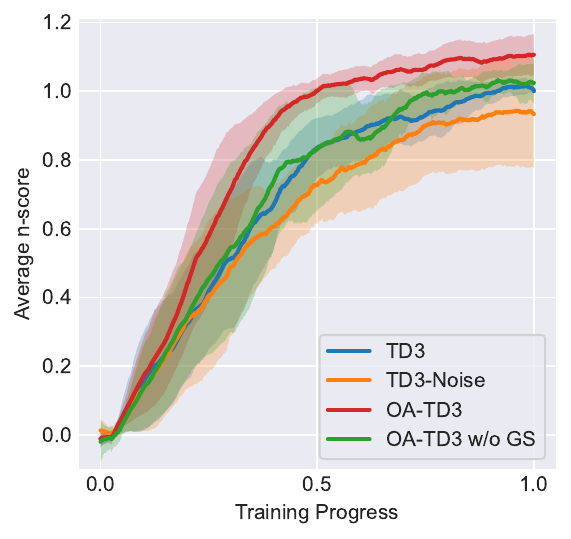}
    \vskip -0.1cm
    \caption{Normalized training curves of OA-TD3 and baseline methods.
    See Fig.~\ref{fig:learning_curves_td3} in Appendix~\ref{sec:app_addtional_res_td3} for individual training curves in each task. }
    \label{fig:learning_curve_td3_all_norm}
    \vskip -1.5cm
\end{wrapfigure}

In this work, we implement OA-TD3 based on TD3 implemented by \cite{huang2022cleanrl}.
The $\epsilon$ value denotes the perturbation strength during robust training, which is tuned in $[0.1,0.3]$ according to different tasks.
As described in Section~\ref{sec:oa_td3}, the hyper-parameter $\omega$ is tuned in $[0.4, 0.6]$ to take a balance between nominal performance and robustness given different tasks.
All methods are trained with at least 10 random seeds.
More details are given in Appendix~\ref{sec:app_exp_details_deterministic}.

\subsubsection{Experiment Results}

The experiment results on robustness are given in Table~\ref{tab:rob_res_td3}.
As demonstrated in the table, OA-TD3 outperforms the baseline methods across both nominal cases and different action perturbations in various tasks.

The training curves are illustrated in Fig.~\ref{fig:learning_curve_td3_all_norm}.
Similar to the analysis of OA-PPO, OA-TD3 (red) achieves sample efficiency that is comparable to or even higher than other methods, including standard TD3 (blue).
This result demonstrate that our method can enhance robustness while maintaining or improving sample efficiency concurrently.

In comparison, TD3-Noise generally exhibits superior robustness performance than typical TD3. 
However, its improvement is modest and accompanied by considerable instability, as evidenced by decreased performance in the \emph{Ant} task under \emph{random} and \emph{biggest} attacks.
Additionally, NR/PR-MDP demonstrates excellent robustness against strong adversaries in some tasks, such as Min-Q and Min-OA-Q attacks in \emph{Walker2d}. 
However, NR/PR-MDP exhibits substantial expense of nominal performance and robustness under weaker perturbations, including random and biggest attacks.
This limitation is prevalent in robust RL under observation perturbations, especially for methods trained concurrently with adversarial attackers, constraining their practical applications.

Moreover, as shown in Table~\ref{tab:rob_res_td3} and Fig.~\ref{fig:learning_curve_td3_all_norm}, \emph{OA-TD3 w/o GS} achieves lower performance on robustness and training efficiency compared to \emph{OA-TD3} across various tasks.
This indicates that the gradient conflict impedes policy optimization, resulting in degradation of both robustness and efficiency.
The utilization of Gradient Surgery discussed in Section~\ref{sec:oa_td3} alleviates this issue effectively, demonstrating its necessity in this work.

\subsection{Hyper-parameter Analysis}

\begin{figure}[htbp]
\vskip -0.1in %
\begin{center}
\hspace{-3.8mm}
\subfigure[$\omega$ in \emph{Walker2d}]{
\includegraphics[height=2.85cm]{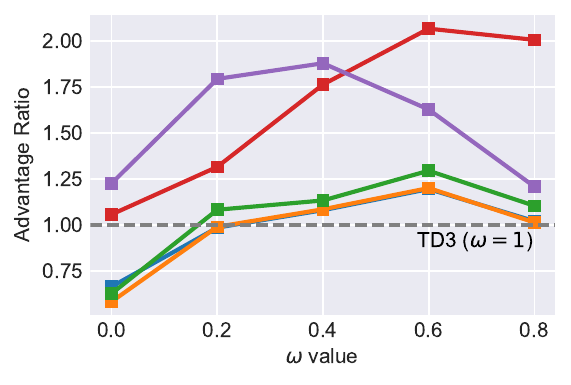}
\label{fig:hyper_omega_walker2d}
}
\hspace{-3.8mm}
\subfigure[$\omega$ in \emph{HalfCheetah}]{
\includegraphics[height=2.85cm]{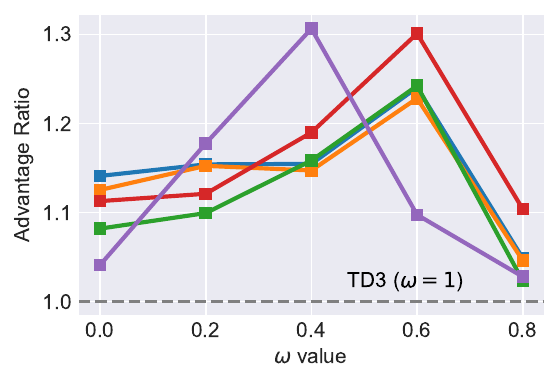}
\label{fig:hyper_omega_halfCheetah}
}
\hspace{-3.8mm}
\subfigure[$\epsilon$  in \emph{Walker2d}]{
\includegraphics[height=2.85cm]{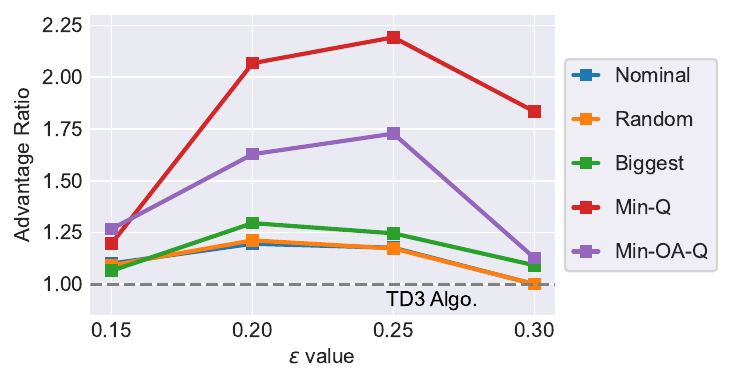}
\label{fig:hyper_epsilon_train_walker2d}
}
\end{center}
\caption{
The influence of parameters $\omega$ and $\epsilon$ to the performance of OA-TD3 policies. 
The $y$-axis denotes the advantage compared to typical TD3 under same adversaries.
Advantage ratio greater than 1.0 means higher performance than TD3.
(a)-(b) Robustness of policies trained with different $\omega$ in \emph{Walker2d} and \emph{HalfCheetah} tasks. 
(c) Robustness of policies trained with different $\epsilon$ in the \emph{Walker2d} task. Policies are evaluated against perturbation strength $\epsilon=0.2$.
}
\label{fig:hyperparameter_analysis}
\end{figure}

In order to analyze the influence of hyper-parameters, experiments with different $\omega$ and $\epsilon$ values are conducted.
As shown in Fig.~\ref{fig:hyper_omega_walker2d} and \ref{fig:hyper_omega_halfCheetah}, the best robustness performance is achieved with $\omega$ around 0.4 and 0.6.
As discussed in Section~\ref{sec:method_OA_DRL},  a larger $\omega$ such as $0.6$ prioritizes the standard critic in policy optimization, resulting in improved performance against weak adversaries, such as random and biggest attacks.
In contrast, $\omega=0.4$ emphasizes OA-aware critic more heavily, leading to better robustness under strong perturbations, such as Min-OA-Q attacks.

As shown in Fig.~\ref{fig:hyper_epsilon_train_walker2d}, policies are trained with different $\epsilon$ values and evaluated against adversaries with strength $0.2$.
Our method outperforms TD3 generally, where $\epsilon$ between 0.20 and 0.25 achieves the best robustness performance.
Optimal robustness under each adversary is achieved under different $\epsilon$ values. 
Smaller $\epsilon$ such as 0.2 emphasizes optimality, corresponding to higher performance under weak attackers, such as random and biggest attackers.
Larger $\epsilon$ values prioritize robustness, leading to higher performance under strong adversaries, such as Min-Q and Min-OA-Q.

\section{Conclusion}
In this work, we formulate the robust RL tasks under action adversaries as AA-MDP problem.
We propose optimal adversary-aware Policy Iteration as a generic framework to enhance policy robustness against action perturbations, which evaluates and improves the performance of the policy under the corresponding optimal adversary.
Our method can be incorporated into mainstream DRL algorithms such as TD3 and PPO, enhancing action robustness while maintaining nominal performance and sample efficiency simultaneously.
Experiment results on various continuous control tasks demonstrate that our method can enhance robustness against different action perturbations effectively.

\bibliography{bibfile.bib}
\bibliographystyle{plain}

\appendix

\section{Limitations and Broader Impacts}
\subsection{Limitations}
\label{sec:app_limitations}
This work mainly studies action robustness of RL policies in continuous control tasks, where policy robustness in discrete decision scenarios are not incorporated, which is one direction for the future research.
Besides, we assume that the perturbation at each step is constrained by the constant strength, which remains fixed throughout the entire episode but may vary in real-world applications.
We will improve AA-MDP formulations to incorporate this issue in future works.

\subsection{Broader Impacts}
\label{sec:app_broader_impacts}
Despite the rapid advancement of RL agents, the performance of well-trained RL policies may be vulnerable to various types of slight perturbations, which may cause catastrophic failure and safety risks.
This paper aims to study and enhance the robustness of RL policies under action perturbations.
This method may be able to advance the deployment of RL in real world applications, especially in safety critical scenarios such as robotic control and autonomous driving.
Besides, this work reveals the vulnerability of DRL policies to slight action perturbations, which may cause potential malicious, such as adversarial attacks in applications.

\section{AA-MDP from the perspective of Adversary}
\label{sec:app_MDP_adversary}

\begin{wrapfigure}{R}{0.53\textwidth}
    \centering
    \vskip -0.35cm
    \includegraphics[width=0.53\textwidth]{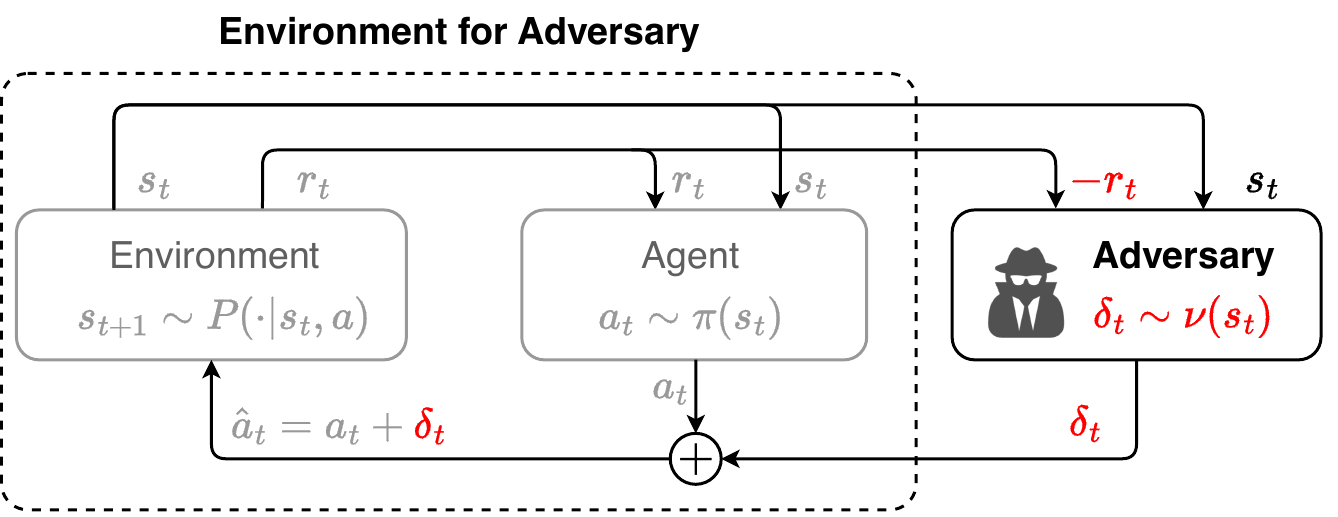}
    \vskip -0.05cm
    \caption{The interaction process of AA-MDP from the perspective of the action adversary, which considers the agent policy $\pi$ as part of the environment.
    }
    \label{fig:interaction_process_adversary}
    \vskip -0.5cm
\end{wrapfigure}

As depicted in Fig.~\ref{fig:interaction_process_adversary}, given a fixed policy $\pi$, the adversary $\nu$ is trained to minimize the performance of the policy $\pi$ in AA-MDP.
This task is equivalent to learning the optimal policy in another MDP $\widetilde{\mathcal{M}}_{\nu} = < \mathcal{S}_{\nu}, \mathcal{A}_{\nu},  P_{\nu}, R_{\nu}, \gamma >$, where  $\mathcal{S}_{\nu} = \mathcal{S}$, $\mathcal{A}_{\nu} = [-\epsilon, \epsilon]^{|\mathcal{A}|}$, $P_{\nu}(s'|s,\delta) = \mathbb{E}_{a\sim\pi}[ P(s'|s,a+\delta) ]$, $R_{\nu}(s,\delta) = - \mathbb{E}_{a\sim\pi}[ R(s,a+\delta) ]$, and $\gamma\in [0,1)$ denotes the discount factor.
Given a fixed policy $\pi$, The objective of the adversary is to minimize the cumulative reward obtained by the agent, i.e.
\begin{equation*}
\begin{aligned}
    \nu^* \leftarrow \arg\max_\nu \mathbb{E}_{s_0\sim \rho, \delta_t\sim \nu, s_{t+1}\sim  P_{\nu}} \left[ \sum_{t=0}^{\infty} \gamma^t R_{\nu}(s_t, \delta_t) \right] 
     = \arg\min_\nu \mathbb{E}_{s} \left[ V_{\pi\circ\nu}(s) \right], \; \text{s.t.} \; \|\delta_t\|_{\infty} \leq \epsilon.
\end{aligned}
\end{equation*}
Empirically, this problem can be solved with RL algorithm to find the $\nu^*$ given a fixed policy $\pi$.
The robust policy can be obtained by training the adversary $\nu$ and the policy $\pi$ iteratively, similar to other robust RL methods that consider different robustness criteria, including PR/NR-MDP~\cite{tessler2019action} for action adversaries and ATLA~\cite{zhang2021robust} for observation adversaries.
However, recent studies~\cite{liang2022efficient} have revealed out that training RL adversaries requires extra samples from the environment, which is even more difficult and sample expensive to solve than the original RL tasks.

\section{Theorems and Proofs}
\label{sec:app_theorem_proofs}
\subsection{Proof of Theorem~\ref{thm:bellman_operator_convergence}}
\label{sec:app_proof_bellman_operator_convergence}
\paragraph{Theorem.} 
Given any fixed policy $\pi \in\Pi$ with $\epsilon\geq 0$, the optimal adversary-aware Bellman operator $\widetilde{\mathcal{T}}^\pi$ is a contraction mapping under sup-norm $\|\cdot\|_{\infty}$, with the fixed point $Q_{\pi\circ\nu^*}$, i.e. $\lim_{k\to\infty}(\widetilde{\mathcal{T}}^\pi)^k Q_{\operatorname{adv}} = Q_{\pi\circ\nu^*}$.
\begin{proof}
The proof is quite similar to the convergence proof of typical Bellman operator $\mathcal{T}^\pi$~\cite{sutton2018reinforcement} and the proof given in WocaR-RL~\cite{liang2022efficient}.
The proof is composed of the following two steps.

Step 1: prove that the optimal adversary-aware Bellman operator $\widetilde{\mathcal{T}}^\pi$ is a contraction mapping.
Given two $Q_{adv}$ functions $Q_1(s,a)$ and $Q_2(s,a)$, we can obtain that:
\begin{equation*}
\begin{aligned}
& \| \widetilde{\mathcal{T}}^{\pi}Q_{1} - \widetilde{\mathcal{T}}^{\pi}Q_{2} \|_{\infty} \\
&= \max_{s,a}\left| R(s,a) + \gamma \mathbb{E}_{P} \min_{\|\delta\|\leq \epsilon} \mathbb{E}_{\pi}\left[ Q_1\left(s',a'+\delta\right) \right] - R(s,a) - \gamma \mathbb{E}_{P} \min_{\|\delta\|\leq \epsilon} \mathbb{E}_{\pi}\left[ Q_2\left(s',a'+\delta\right) \right] \right| \\
&= \gamma \max_{s,a}\left| \mathbb{E}_{s'\sim P}\left[ \min_{\|\delta\|\leq \epsilon} \mathbb{E}_{\pi}\left[ Q_1\left(s',a'+\delta\right) \right] - \min_{\|\delta\|\leq \epsilon} \mathbb{E}_{\pi}\left[ Q_2\left(s',a'+\delta\right) \right] \right] \right| \\
&\leq \gamma \max_{s,a}\mathbb{E}_{s'\sim P}\left| \min_{\|\delta\|\leq \epsilon} \mathbb{E}_{\pi}\left[ Q_1\left(s',a'+\delta\right) \right] - \min_{\|\delta\|\leq \epsilon} \mathbb{E}_{\pi}\left[ Q_2\left(s',a'+\delta\right) \right] \right| \\
&\leq \gamma \max_{s,a}\mathbb{E}_{s'\sim P} \max_{\|\delta\|\leq \epsilon}\big| \mathbb{E}_{\pi}\left[ Q_1\left(s',a'+\delta\right) - Q_2\left(s',a'+\delta\right) \right] \big| \\
&\leq \gamma \max_{s,a}\mathbb{E}_{P,\pi} \left[ \max_{\|\delta\|\leq \epsilon}\big|  Q_1\left(s',a'+\delta\right)  -  Q_2\left(s',a'+\delta\right) \big| \right] \\
&= \gamma \max_{s,a}\mathbb{E}_{P,\pi} \left[ \left\| Q_1  -  Q_2 \right\|_{\infty} \right] \\
&= \gamma \left\| Q_1 -  Q_2 \right\|_{\infty}. \\
\end{aligned}
\end{equation*}
Thus, $\widetilde{\mathcal{T}}^\pi$ is a contraction mapping under the sup-norm $\| \cdot \|_{\infty}$.

Step 2: $Q_{\pi\circ\nu^*}$ is the fixed point.
Assuming that $Q_{adv}$ converges to $Q^*$, we can obtain the following equation based on the definition of $\widetilde{\mathcal{T}}^\pi$.
Given any $s_0\in \mathcal{S}$ and $a_0\in \mathcal{A}$, we have:
\begin{equation*}
\begin{aligned}
    Q^*(s_0,a_0) 
    &= \left(\widetilde{\mathcal{T}}^\pi Q^*\right)(s_0,a_0)\\
    &= R(s_0,a_0) + \gamma \mathbb{E}_{s_1\sim P} \min_{\|\delta_1\|\leq \epsilon} \mathbb{E}_{a_1\sim \pi}\left[Q^*(s_1,a_1+\delta_1) \right] \\
    &= R(s_0,a_0) + \gamma \mathbb{E}_{s_1\sim P} \min_{\|\delta_1\|\leq \epsilon} \mathbb{E}_{a_1\sim \pi}\left[\left(\widetilde{\mathcal{T}}^\pi Q^*\right)(s_1,a_1+\delta_1) \right] \\
    &=  ... \\
    &= R(s_0,a_0) + \gamma \mathbb{E}_{s_i\sim P} \min_{\|\delta_i\|\leq \epsilon} \mathbb{E}_{a_i\sim \pi}\bigg[ \sum_{t=1}^{\infty} R(s_t, a_t + \delta_t) \bigg] \\
    &= R(s_0,a_0) + \gamma \mathbb{E}_{s_1\sim P} \left[ V_{\pi\circ\nu^*}(s_1) \right] \\
    &= Q_{\pi\circ\nu^*} (s_0, a_0).
\end{aligned}
\end{equation*}
Thus, $Q^* = Q_{\pi\circ\nu^*}$ is the fixed point, which evaluates the performance of the policy $\pi$ under the optimal adversary $\nu^*$.
Therefore, we have:
\begin{equation*}
\begin{aligned}
\lim_{k\to\infty} \|  (\widetilde{\mathcal{T}}^{\pi})^k Q_{\operatorname{adv}} - Q_{\pi\circ\nu^*} \|_\infty
&= \lim_{k\to\infty} \|  (\widetilde{\mathcal{T}}^{\pi})^k Q_{\operatorname{adv}} - (\widetilde{\mathcal{T}}^{\pi})^k Q_{\pi\circ\nu^*} \|_\infty  \\
&\leq \lim_{k\to\infty} \gamma^k \|  Q_{\operatorname{adv}} - Q_{\pi\circ\nu^*} \|_\infty \\
& = 0,
\end{aligned}
\end{equation*}
Above all, we can obtain that $\lim_{k\to\infty}(\widetilde{\mathcal{T}}^\pi)^k Q_{\operatorname{adv}} = Q_{\pi\circ\nu^*}$.
\end{proof}

\subsection{Proof of Theorem~\ref{thm:policy_improvement_thm}}
\label{sec:app_policy_improvement_thm}

\paragraph{Theorem.}
Given any two policies $\pi$ and $\pi'$, 
we define $Q_{\operatorname{adv}}^{\pi} (s,\pi') = \min_{\| \delta \| \leq \epsilon} \mathbb{E}_{a\sim\pi'}[ Q_{\operatorname{adv}}^{\pi}(s,a+\delta) ]$.
For $\forall s\in\mathcal{S}$, if $Q_{\operatorname{adv}}^{\pi} (s,\pi')\geq V_{\pi\circ\nu^*}(s)$, then we have $V_{\pi'\circ\nu^*}(s) \geq V_{\pi\circ\nu^*}(s)$ for $\forall s\in\mathcal{S}$, i.e. $\pi'$ is at least good as policy $\pi$.

\begin{proof}
The proof is similar to the proof of Policy Improvement Theorem in the typical MDP $\mathcal{M}$.
We know that $V_{\pi\circ\nu^*}(s_0) \leq Q_{\operatorname{adv}}^{\pi} (s_0,\pi')$ for $\forall s_0\in \mathcal{S}$, which can be expanded as follows:
\begin{equation*}
\begin{aligned}
    V_{\pi\circ\nu^*}(s_0) &\leq Q_{\operatorname{adv}}^{\pi}(s_0,\pi') \\
    &= \min_{\|\delta_0\|\leq\epsilon} \mathbb{E}_{a_0\sim\pi', s_1\sim P}\left[ R\left(s_0, a_0+\delta_0\right) + \gamma V_{\pi\circ\nu^*}(s_1)  \right] \\
    &\leq \min_{\|\delta_i\|\leq\epsilon} \mathbb{E}_{a_0\sim\pi', s_1\sim P}\left[ R\left(s_0, a_0+\delta_0\right) + \gamma Q_{\operatorname{adv}}^{\pi}(s_1,\pi')  \right] \\
    &= \min_{\|\delta_i\|\leq\epsilon} \mathbb{E}_{a_i\sim\pi', s_i\sim P}\left[ R\left(s_0, a_0+\delta_0\right) + \gamma R\left(s_1, a_1+\delta_1\right) + \gamma^2 V_{\pi\circ\nu^*}(s_2)  \right] \\
    &\leq \cdots \\
    & \leq \min_{\|\delta_i\|\leq\epsilon} \mathbb{E}_{a_i\sim\pi', s_i\sim P}\left[ \sum_{i=0}^{\infty} \gamma^i R\left(s_i, a_i+\delta_i\right)  \right] \\
    &= V_{\pi'\circ\nu^*}(s_0)
\end{aligned}
\end{equation*}
\end{proof}

\subsection{Proof of Theorem~\ref{thm:policy_improvement}}
\label{sec:app_proof_policy_improvement}

\paragraph{Theorem.}
Given $\forall \epsilon\geq 0$, $\pi$ and $\pi'$ denote policies before and after optimal adversary-aware policy improvement, we have $V_{\pi'\circ\nu^*}(s)\geq V_{\pi\circ\nu^*}(s)$ for $\forall s\in \mathcal{S}$.

\begin{proof}
This proof can be proved based on Theorem~\ref{thm:policy_improvement_thm}.
As illustrated in Eq.~\eqref{eq:policy_improvement}, the new policy $\pi'$ can be formulated as:
$\pi'\leftarrow \arg\max_{\pi'\in\Pi} \min_{\|\delta\|\leq \epsilon} \mathbb{E}_{a\sim\pi'} \left[ Q_{\operatorname{adv}}(s, a+\delta) \right]$.
Thus, we have: 
\begin{equation*}
Q_{\operatorname{adv}}^{\pi} (s,\pi') = \min_{\|\delta\|\leq \epsilon}\mathbb{E}_{\pi'} Q_{\operatorname{adv}}(s,a+\delta) 
\geq \min_{\|\delta\|\leq \epsilon} \mathbb{E}_{\pi}\left[ Q_{\operatorname{adv}}(s,a+\delta) \right]
= V_{\pi\circ\nu^*}(s),
\end{equation*}
i.e. $Q_{\operatorname{adv}}^{\pi} (s,\pi') \geq V_{\pi\circ\nu^*}(s)$, which is the requirement of Theorem~\ref{thm:policy_improvement_thm}.
Therefore, we can obtain that $\forall s\in\mathcal{S}, V_{\pi'\circ\nu^*}(s)\geq V_{\pi\circ\nu^*}(s)$, i.e.  $\pi'$ is at least good as policy $\pi$.
\end{proof}

\subsection{Proof of Theorem~\ref{thm:policy_iteration_convergence}}
\label{sec:app_proof_policy_iteration_convergence}

\paragraph{Theorem.}
Given $\forall Q_{\operatorname{adv}}, \pi\in\Pi$ for initialization, the policy will converges to the optimal policy $\pi^*$ after sufficient optimal adversary-aware Policy Iterations.

\begin{proof}
We utilize $\pi_k\in\Pi, k\in\mathbb{Z}$ denotes the policy trained with OA-PI algorithm after $k$ iterations.
Theorem~\ref{thm:policy_improvement} have told us that the OA-PI achieves non-decreasing update of the policy, $V_{\pi_{k+1}\circ\nu^*}(s)\geq V_{\pi_k\circ\nu^*}(s)$ for $\forall k\in\mathbb{Z}, s\in\mathcal{S}$.

When $V_{\pi_{k+1}\circ\nu^*}(s)= V_{\pi_k\circ\nu^*}(s)$, the algorithm converges with $\pi_k=\pi_{k+1}$.
We have that
$$
V_{\pi_{k}\circ\nu^*}(s) = Q_{\operatorname{adv}}^{\pi_k} (s,\pi_k) = \max_a \min_{\|\delta\|\leq \epsilon} Q_{\operatorname{adv}}(s,a+\delta).
$$
Thus, we can obtain the following equation based on Eq.~\eqref{eq:bellman_operator}:
\begin{equation*}
\begin{aligned}
    Q_{\operatorname{adv}}(s,a) 
    = R(s,a) + \gamma \mathbb{E}_{P}\left[ V_{\pi_{k}\circ\nu^*}(s) \right] 
    = R(s,a) + \gamma \mathbb{E}_{P}\left[ \min_{\|\delta\|\leq \epsilon} \max_a Q_{\operatorname{adv}}(s,a+\delta) \right],
\end{aligned}
\end{equation*}
which satisfies the Bellman optimality equation in AA-MDP.
Thus, current policy $\pi_k=\pi^*$ is the optimal policy, i.e. the policy iteration converges to the optimal policy.
\end{proof}

\section{Optimal Adversary-Aware DRL Algorithm}

\subsection{Optimal Adversary-Aware PPO Algorithm}
\label{sec:app_oa_ppo_algo}

The pseudo-code of optimal adversary-aware PPO algorithm is given in Algorithm~\ref{alg:oa_ppo}.

\begin{algorithm}[htb]
   \caption{Optimal adversary-aware PPO}
   \label{alg:oa_ppo}
\begin{algorithmic}[1]
   \STATE {\bfseries Input:} adversary strength $\epsilon\geq 0$, weight $\omega\in [0,1]$, number of iterations $T$.
   \STATE Initialize policy network $\pi$, value network $V(s)$, and OA-aware critic $Q_{\operatorname{adv}}$.
   \FOR{$k=1$ \textbf{to} $T$}
    \STATE Collect a set of trajectories $\mathcal{D}=\{\tau_i\}$ by running policy $\pi$ in the environment, where $\tau_i=\{ s_t, a_t, r_t, s_{t+1}, d_t \}$.
    \STATE Compute rewards-to-go $\hat{R}_t$ for each step $t$ in each trajectory.
    \STATE Compute advantage estimation $\hat{A}_t$ based on the current value function $V(s)$ and $\hat{R}_t$.
    \STATE Update the value function $V(s)$ by regression on mean-squared error: 
    \[\theta^V \leftarrow \arg\min_{\theta^V} \frac{1}{\left|\mathcal{D}\right|\left|\tau_i\right|} 
    \sum_{\tau_i\in\mathcal{D}} 
    \sum_{t=0}^{\left|\tau_i\right|}
    \left( V\left(s_t\right)-\hat{R}_t \right)^2\]
    \STATE Compute $y_{\operatorname{adv}}$ on trajectories $\tau_i\in \mathcal{D}$:
    $
    y_{t} = r_t + \gamma (1-d_t) \min_{\|\delta\|<\epsilon} Q_{\operatorname{adv}}(s_{t+1}+\delta)
    $.
    \STATE Update the OA-aware critic $Q_{\operatorname{adv}}$:
    \[\theta^{Q_{\operatorname{adv}}} \leftarrow \arg\min_{\theta^{Q_{\operatorname{adv}}}} \frac{1}{\left|\mathcal{D}\right|\left|\tau_i\right|} 
    \sum_{\tau_i\in\mathcal{D}} 
    \sum_{t=0}^{\left|\tau_i\right|}
    \left( y_t - Q_{\operatorname{adv}}(s_t, a_t) \right)^2\]
    \STATE Update the policy $\pi$:
    \[
    \theta^{\pi} \leftarrow \arg\min_{\theta^{\pi}} \frac{1}{\left|\mathcal{D}\right|\left|\tau_i\right|} 
    \sum_{\tau_i\in\mathcal{D}} 
    \sum_{t=0}^{\left|\tau_i\right|}
    \left[  \min\left( \rho_{\pi} \widetilde{A}(s,a), \, \operatorname{clip}(\rho_{\pi}, 1-\xi, 1+\xi) \widetilde{A}(s,a) \right) \right],
    \]
    \quad\; where $\rho_{\pi}=\frac{\pi(a|s)}{\pi_{\operatorname{old}}(a|s)}$ and $\widetilde{A}_(s,a) = \omega A(s,a) + (1-\omega) Q_{\operatorname{adv}}(s,a+\delta^*)$.
   \ENDFOR
\end{algorithmic}
\end{algorithm}

\subsection{Gradient Conflict in OA-TD3 Algorithm}
\label{sec:app_pcgrad}

As described in Section~\ref{sec:oa_td3}, the policy $\mu$ in OA-TD3 is trained with typical $Q$ and OA-aware $Q_{\operatorname{adv}}$, which are formulated as follows:
\begin{equation*}
\mu \leftarrow\max_{\mu}\mathbb{E}_{s\sim\mathcal{D}}\left[\omega Q(s,\mu(s)) + (1-\omega) Q_{\operatorname{adv}}(s,\mu(s)+\delta^*) \right].
\end{equation*}
The policy $\mu$ is trained utilizing two gradients from two critics $\nabla_{\theta^{\mu}}Q)$ and $\nabla_{\theta^{\mu}}Q_{\operatorname{adv}}$, corresponding to optimality and robustness accordingly.
Intuitively, these two objectives are contradictory to some extent, i.e. optimality and robustness are hard or even unable to be achieved simultaneously.
In this work, we employ the definition of gradient conflict~\cite{yu2020gradient} proposed for multi-task learning to formalize this problem.
The definition is illustrated as follows:
\begin{definition}
    Given any two gradients $\mathbf{g}_i$ and $\mathbf{g}_j$, $\phi_{i,j}$ denotes the angle between two gradients, we define the gradients as conflicting when $\cos{\phi_{i,j}}<0$. %
\end{definition}

OA-TD3 may experience conflicting gradients during policy updates when $(\nabla_{\theta^{\mu}}Q)\cdot (\nabla_{\theta^{\mu}}Q_{\operatorname{adv}}) < 0$.
Such conflicting gradients impede the optimization process for both two objectives, leading to unstable and inefficient training, and potentially degrading policy performance.

\begin{figure}[h]
\begin{center}
\centerline{\includegraphics[width=0.98\columnwidth]{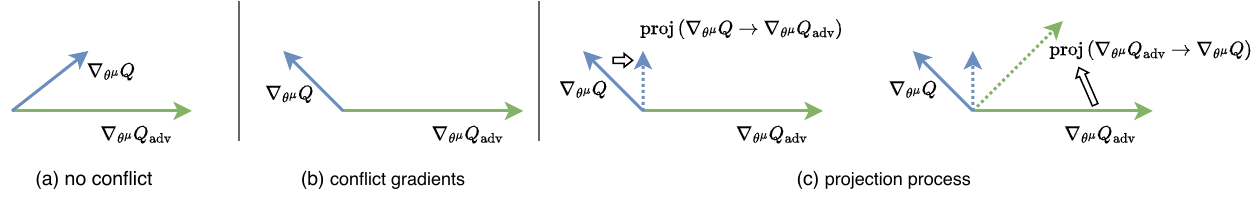}}
\caption{
The process of projection operations in Gradient Surgery algorithm.
}
\label{fig:pcgrad}
\end{center}
\end{figure}
In order to solve the gradient conflicting problem, we employ the Projecting Conflicting Gradients (PCGrad)~\cite{yu2020gradient} in this work.
PCGrad is a popular implementation of Gradient Surgery (GS) and commonly utilized in multi-task learning problems.
As illustrated in Fig.~\ref{fig:pcgrad}, when the gradient conflicts are observed, we modify the direction and magnitude of gradients by projecting each gradient onto the normal plane of the other.
This method is computationally efficient and resolves conflicts effectively.
The experiment results in Section~\ref{sec:experiment_oa_td3} demonstrate that GS is effective and necessary for robustness performance and sample efficiency.
Note that the gradient conflicts only exist in OA-aware training in deep deterministic policy gradient, such as OA-TD3.
Thus, the Gradient Surgery is unnecessary for other RL algorithms, such as OA-PPO algorithm. 

\section{Experiment Details}
\label{sec:app_experiment_details}
\subsection{Computation Resources}
\label{sec:computation_resources}
\begin{wraptable}{r}{0.6\textwidth}
\vskip -0.25in
\caption{
The time cost for training each method in \emph{Walker2d} and \emph{Ant} tasks.
OA-PPO (SS) means OA-PPO is trained with same steps as PPO.
OA-PPO (SP) means OA-PPO is trained to achieve the same nominal performance to PPO algorithm.
}
\label{tab:time_cost}
\vskip -0.05in
\begin{small}
\begin{center}
\begin{tabular}{lccc}
\toprule
Task & PPO & OA-PPO (SS) & OA-PPO (SP)\\ \midrule
Walker2d & 1.60h & 3.05h & 1.48h (978K steps) \\  
Ant & 1.75h & 3.20h & 2.35h (1.48M steps) \\  
\bottomrule  
\toprule
Task & TD3 & OA-TD3 (SS) & OA-TD3 (SP)\\ \midrule
Walker2d & 1.87h & 5.41h & 2.70h (501K steps) \\  
Hopper & 2.10h & 5.48h & 2.31h (425K steps) \\  
HalfCheetah & 2.10h & 5.52h & 2.35h (429K steps) \\  
\bottomrule
\end{tabular} 
\end{center}
\end{small}
\vskip -0.1in
\end{wraptable}
In this work, experiments are conducted with AMD EPYC 7763 CPU, NVIDIA RTX 3090 GPU, and 256G memory.
The training time of our method compared to typical DRL is shown in Table~\ref{tab:time_cost}.
Generally, our method requires more training times than vanilla DRL.
Most additional times are utilized to find $\delta^*$ by the gradient descent operations of PGD, which are computationally expensive and hard to parallelized. 
Fortunately, OA-DRL is more sample efficient than vanilla DRL, which needs less training steps to achieve same nominal performance. 
We are exploring alternative methods to find $\delta^*$, such as Genetic Algorithm and simulated annealing, as potential replacements for PGD to reduce computational costs.

\subsection{Experiment Details for OA-PPO}
\label{sec:app_exp_details_stochastic}

\begin{table}[t]
\begin{center}
\parbox{.48\linewidth}{
\centering
\label{tab:hyperparameter_ppo}
\caption{The default hyper-parameter settings for PPO-based methods  in the experiments.}
\begin{tabular}{lc}
\toprule
Parameter & Setting\\
\midrule
Learning rate & $3\times 10^{-4}$\\
Optimizer  & Adam\\
Learning rate decay & Linear\\
Total timestep & $1\times 10^{6}$ \\
Discount factor $\gamma$ & 0.99\\
GAE $\lambda$ value & 0.95\\
Rollout steps & 2048\\
Number of mini-batches & 32\\
Batch size & 64 \\
Update epochs & 10\\
Surrogate clipping & 0.2\\
Gradient norm clipping & 0.5\\
Perturbation strength $\epsilon$ & [0.1, 0.3] \\
Policy improvement weight $\omega$ & [0.4, 0.6] \\
\bottomrule
\end{tabular}
}
\hspace{10pt}
\parbox{.48\linewidth}{
\centering
\label{tab:hyperparameter_td3}
\caption{The default hyper-parameter settings for TD3-based methods in the experiments.}
\begin{tabular}{lc}
\toprule
Parameter & Setting\\
\midrule
Learning rate  & $3\times 10^{-4}$\\
Optimizer  & Adam\\
Total timestep & $1\times 10^{6}$ \\
Buffer size &  $1\times 10^{6}$ \\
Soft update $\tau$ & 0.005 \\
Batch size  & 256 \\
Exploration noise $\sigma$  & 0.1 \\
Learning starts timestep  & 25000 \\
Policy update delay & 2 \\
Policy smoothing noise $\sigma$ & 0.2 \\
Policy smoothing clip & 0.5 \\
Discount factor $\gamma$ & 0.99\\
Perturbation strength $\epsilon$ & [0.1, 0.3] \\
Policy improvement weight $\omega$ & [0.4, 0.6] \\
\bottomrule
\end{tabular}
}
\end{center}
\end{table}

\begin{table}[t]
\caption{ The $\epsilon$ values for each environment. Note that the action space of each environment is normalized to $[-1, 1]^{|A|}$ for simplicity.
}
\begin{small}
\begin{center}
\begin{tabular}{lccclccc}
\toprule
Task & Obs. Space & Action Space & $\epsilon$  & Task & Obs. Space & Action Space & $\epsilon$  \\ \midrule
Ant & $\mathbb{R}^{27}$ & $[-1,1]^{8}$ & 0.15  &  LunarLander & $\mathbb{R}^{8}$ & $[-1,1]^{2}$ & 0.3 \\  
Hopper & $\mathbb{R}^{11}$ & $[-1,1]^{3}$ & 0.2 &  HalfCheetah & $\mathbb{R}^{17}$ & $[-1,1]^{6}$ & 0.2 \\  
Walker2d & $\mathbb{R}^{17}$ & $[-1,1]^{6}$ & 0.2 &  BipedalWalker & $\mathbb{R}^{24}$ & $[-1,1]^{4}$ & 0.2 \\  
\bottomrule  
\end{tabular} 
\end{center}
\end{small}
\end{table}

\paragraph{PPO}
In this work, we utilize PPO implementation proposed by CleanRL~\cite{huang2022cleanrl}.
The hyper-parameters are shown in Table~\ref{tab:hyperparameter_ppo}, which are utilized in most tasks in this work.

\paragraph{PPO-Noise}
Different to typical PPO, PPO-Noise is trained in polluted environment, where the agent takes polluted actions $\hat{a}_t = a_t + \delta$ with random action noises $\delta\sim U[-\epsilon, \epsilon]$ at each step.
The $\epsilon$ denotes the strength of the action perturbation.
Similar to TD3-Noise, $\epsilon$ is chosen in $[0.05, 0.20]$ to take a balance between nominal performance and robustness.

\paragraph{OA-PPO}
In this work, we implement OA-PPO following Algorithm~\ref{alg:oa_ppo} based on PPO implemented by CleanRL~\cite{huang2022cleanrl}.
Most settings for OA-PPO are same to that in the typical PPO.
The perturbation strength $\epsilon$ is selected within the range of 0.1 to 0.3, while the weight $\omega$ is chosen within the range of 0.4 to 0.6. 
These parameter values are adjusted to strike a balance between achieving optimality and robustness, as they may vary depending on the specific environment.

During  policy evaluation and policy improvement of OA-TD3, we utilize the Projected Gradient Decent (PGD) attack~\cite{madry2018towards} to find the optimal action perturbations $\delta^*$ based on the Gradient Decent.
Take the policy improvement of OA-TD3 as an example, we need to find $\delta^* = \arg\min_{\|\delta\|\leq \epsilon} Q_{\operatorname{adv}}(s,\mu(s)+\delta)$, which can be obtained as follows:
\begin{equation*}
    {\delta}^{k+1} = \operatorname{CLIP}\left( {\delta}^k + \eta\cdot \operatorname{sign}\left( \nabla_{{\delta}^k} Q_{\operatorname{adv}}(s, \mu(s) + \delta^k) \right), -\epsilon, \epsilon \right), 
    \quad 0\leq k < K,
\end{equation*}
where $\eta$ denotes step size, $k$ denotes the step count, and we have $\delta^* \leftarrow \delta^K$.
In this work, we set $K$ as $15\sim 20$ and $\eta = \epsilon/K$.

\paragraph{PPO-Min-Q}
PPO-Min-Q is similar to OA-PPO, which replace OA-aware critic $Q_{\operatorname{adv}}$ in OA-PPO with typical Q functions.
Thus, the policy $\pi$ in PPO-Min-Q is trained as follows:
\begin{equation*} 
{\pi}\leftarrow \arg
\max_{\pi} 
\mathbb{E}_{(s,a)\sim\mathcal{D}}\left[ 
\min\left( 
\rho_{\pi} \widetilde{A}(s,a), \,
g(\rho_{\pi}) \widetilde{A}(s,a) 
\right)
\right],
\end{equation*}
where $\widetilde{A}_(s,a) = \omega A(s,a) + (1-\omega) Q(s,a+\delta^*)$ and $\delta^* = \arg\min_{\|\delta\|\leq \epsilon} Q_{\operatorname{adv}}(s,a+\delta)$ denotes the action perturbations.
Other settings are same to OA-PPO algorithm.
Experiments of PPO-Min-Q mainly aims to demonstrate the necessity and effectiveness of OA-aware critic in OA-PPO algorithm.

\subsection{Experiment Details for OA-TD3}
\label{sec:app_exp_details_deterministic}

\paragraph{TD3} 
In this work, we utilize TD3 implementation proposed by CleanRL~\cite{huang2022cleanrl}, which provides a high-quality implementation of DRL algorithms, including TD3 and PPO utilized in this work.
The default hyper-parameters are shown in Table~\ref{tab:hyperparameter_td3}, which are utilized in most tasks in this work.
Modifications are made based on this settings in some special tasks, such as setting the learning rate as 0.001 and the total steps as $5\times 10^5$ in \emph{LunarLander}.

\paragraph{TD3-Noise}
Different to typical TD3, TD3-Noise is trained in polluted environment, where the agent takes polluted actions $\hat{a}_t = a_t + \delta$ with random action noises $\delta\sim U[-\epsilon, \epsilon]$ at each step.

\paragraph{NR/PR-MDP}
NR/PR-MDP~\cite{tessler2019action} denote two RL methods to enhance action robustness against action adversaries in Probabilistic Action Robust MDP (PR-MDP) and Noisy Action Robust MDP (NR-MDP) respectively.
\textbf{(1) PR-MDP:} the agent takes polluted actions with a fixed probability.
\textbf{(2) NR-MDP:} the agent takes actions interpolated between clean actions and polluted actions.
Both two methods enhance policy robustness with adversarial training, i.e. training policies alongside adversarial action attackers iteratively.
In this experiment, we train robust policies utilizing two methods with hyper-parameters recommended in the original paper. 
The higher score of two baseline methods are reported for each task and adversary.

\paragraph{OA-TD3}
In this work, we implement OA-TD3 following Algorithm~\ref{alg:oa_td3} based on TD3 implemented by CleanRL~\cite{huang2022cleanrl}.
Most settings for OA-TD3 are same to that in TD3.
The perturbation strength $\epsilon$ is selected within the range of 0.1 to 0.3, while the weight $\omega$ is chosen within the range of 0.4 to 0.6. 
These parameter values are adjusted to strike a balance between achieving optimality and robustness, as they may vary depending on the specific environment.

\paragraph{OA-TD3 w/o GS}
This is an ablation study to demonstrate the effectiveness of Gradient Surgery in the OA-TD3 algorithm.
The Gradient Surgery is removed, while other settings are primarily the same as those of OA-TD3.

\subsection{Min-OA-Q Attacks}
\label{sec:app_min_oa_q_attacks}

\begin{algorithm}[htb]
   \caption{Train optimal adversary-aware Q function (OA-Q) for Min-OA-Q attacks}
   \label{alg:train_oa_q}
\begin{algorithmic}
   \STATE {\bfseries Input:} Policy $\mu$ to be attacked, adversary strength $\epsilon\geq 0$.
   \STATE {\bfseries Output:} Well-trained OA-Q function $Q_{\operatorname{adv}}$.
   \STATE Initialize OA-Q function $Q_{\operatorname{adv}}$ and $Q_{\operatorname{adv}}'$ with weights $\theta^{Q_{\operatorname{adv}}'}\leftarrow\theta^{Q_{\operatorname{adv}}}$.
   \STATE Initialize replay buffer $\mathcal{D}$ through exploration with $a_t = \mu(s_t) + \xi$, where $\xi\sim\mathcal{N}$.
   \FOR{$t=1$ \textbf{to} $T$}
    \STATE Select action $a_t = \mu(s_t) + \delta^*$, where $\delta^* = \arg\min_{\|\delta\|\leq \epsilon} Q_{\operatorname{adv}}(s,\mu(s)+\delta)$
    \STATE Execute $a_t$ and observe $s_{t+1}$ with reward $r_t$.
    \STATE Store transition $(s_t, a_t, r_t, s_{t+1},d_t)$ in buffer $\mathcal{D}$
    \STATE Sample a batch of data $B=\left\{ \left( s,a,r,s',d \right) \right\}$ from $\mathcal{D}$
    \STATE Compute $a'= \mu'(s')+\xi$, where $\xi\sim\operatorname{clip}\left(\mathcal{N},-c,c \right)$
    \STATE $y_{\operatorname{adv}} = r+\gamma (1-d)\min_{\|\delta\|\leq \epsilon} Q_{\operatorname{adv}}'(s',a'+\delta)$
    \STATE $\theta^{Q_{\operatorname{adv}}}\leftarrow \arg\min \mathbb{E}_{(s,a)\in B} (y_{\operatorname{adv}} - Q_{\operatorname{adv}}(s,a))$
    \STATE Update target networks $Q_{\operatorname{adv}}'$ using soft updates.
   \ENDFOR
\end{algorithmic}
\end{algorithm}

Min-OA-Q attacks means adding action perturbations to minimize the OA-aware Q values $Q_{\operatorname{adv}}$ of the given policy.
Given a fixed policy $\mu$at state $s_t$, the agent take polluted action $\hat{a}_t$ illustrated as follows:
$$
\hat{a}_t = \mu(s_t) + \delta, \; \text{where} \; \delta = \arg\min_{\|\delta\|\leq \epsilon} Q_{\operatorname{adv}}(s_t,\mu(s)+\delta).
$$
In this work, the action perturbation $\delta$ is obtained through $30$ steps PGD attacks.
In order to implement Min-OA-Q attacks on any policy $\mu$ with any perturbation strength $\epsilon>0$, we need to train $Q_{\operatorname{adv}}$ alone.
The process is illustrated in Algorithm~\ref{alg:train_oa_q}.

\subsection{Normalized Nominal Performance}
\label{sec:app_normalize_reward}

Different tasks can have vastly different reward scales, thus cannot be directly averaged.
In order to depict the average episode return during training across various tasks, we compute the normalized normalized score (n-score) of each method, which is widelt utilized in prior works~\cite{Hessel2017RainbowCI,Yu2021TAACTA}.
Given an episode return $Z$, its n-score is calculated as $Z_{norm} = \frac{Z-Z_0}{Z_1-Z_0}$, where 
$Z_0$ and $Z_1$ denote the episode return of random policies and vanilla DRL policies respectively.
The n-score $Z_{norm}\in [0,1]$, thus can be averaged across different tasks.

\section{Additional Experiment Results}
\label{sec:app_additional_experiment_res}

\subsection{Additional Results of OA-PPO}
\label{sec:app_addtional_res_ppo}

\begin{figure}[htbp]
\begin{center}
\hspace{-4.1mm}
\subfigure[\emph{Bipedalwalker}]{
\includegraphics[width=0.249\linewidth]{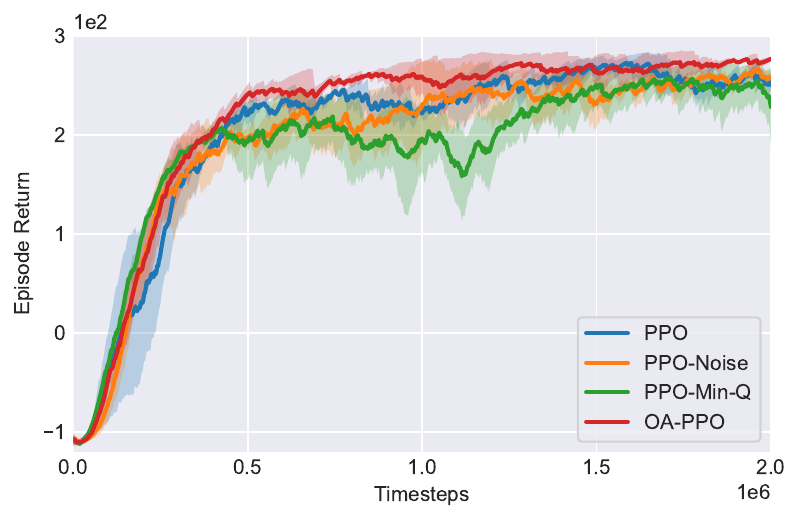}
}
\hspace{-4.1mm}
\subfigure[\emph{Walker2d}]{
\includegraphics[width=0.249\linewidth]{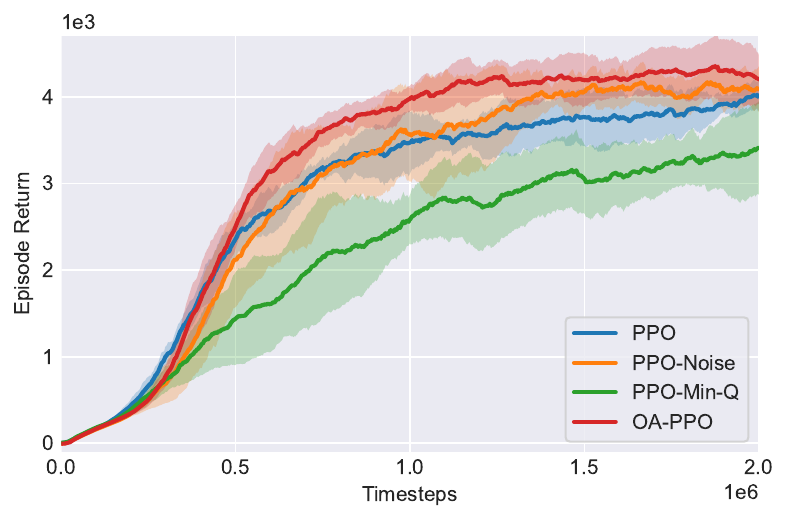}
}
\hspace{-4.1mm}
\subfigure[\emph{Ant}]{
\includegraphics[width=0.249\linewidth]{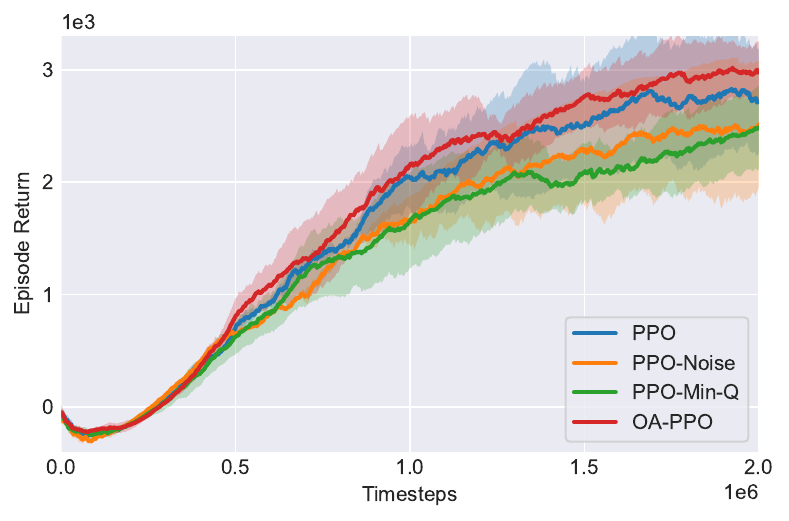}
}
\hspace{-4.1mm}
\subfigure[\emph{Hopper}]{
\includegraphics[width=0.249\linewidth]{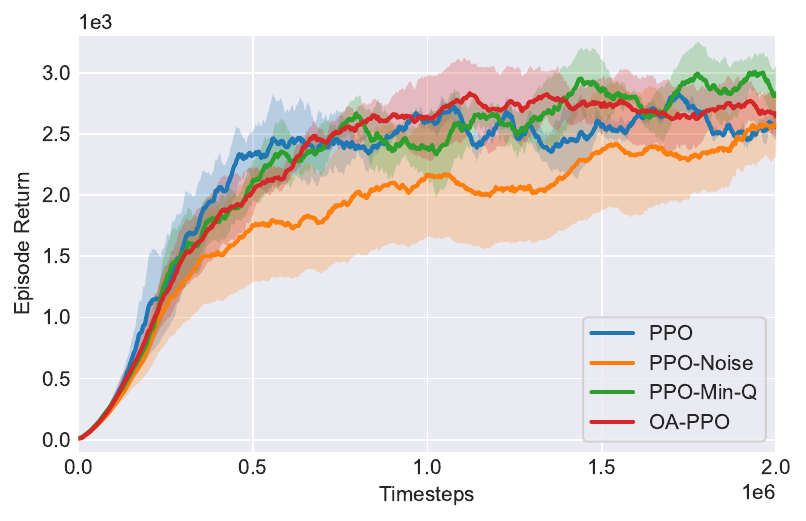}
}
\end{center}
\caption{
Learning curves of OA-PPO (red) in various tasks against baseline methods.
}
\label{fig:learning_curves_ppo}
\end{figure}

\subsection{Additional Results of OA-TD3}
\label{sec:app_addtional_res_td3}

\begin{figure}[htbp]
\begin{center}
\hspace{-4.1mm}
\subfigure[\emph{Walker2d}]{
\includegraphics[width=0.249\linewidth]{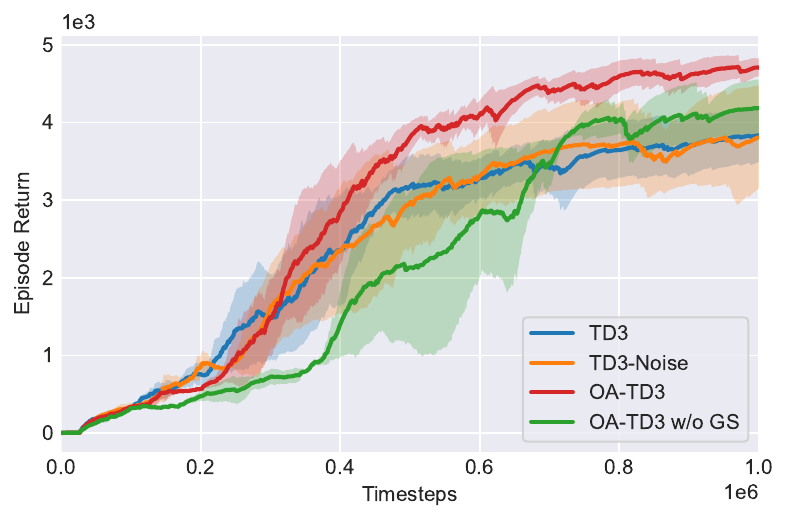}
}
\hspace{-4.1mm}
\subfigure[\emph{Hopper}]{
\includegraphics[width=0.249\linewidth]{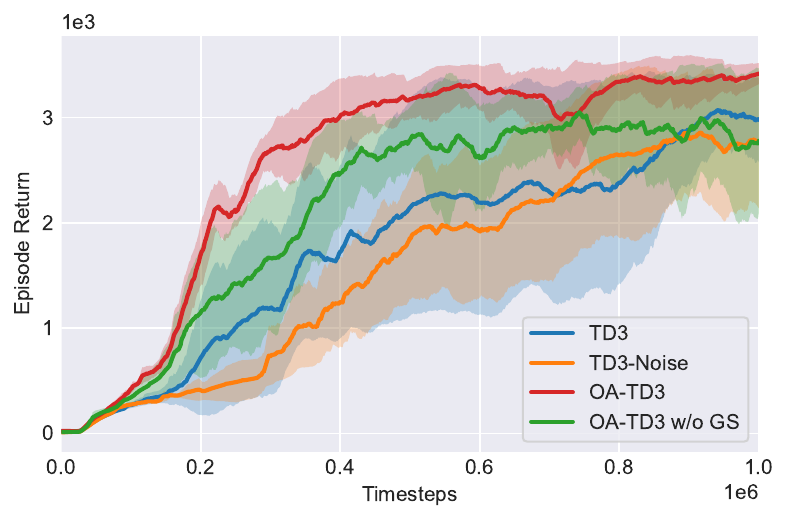}
}
\hspace{-4.1mm}
\subfigure[\emph{Humanoid}]{
\includegraphics[width=0.249\linewidth]{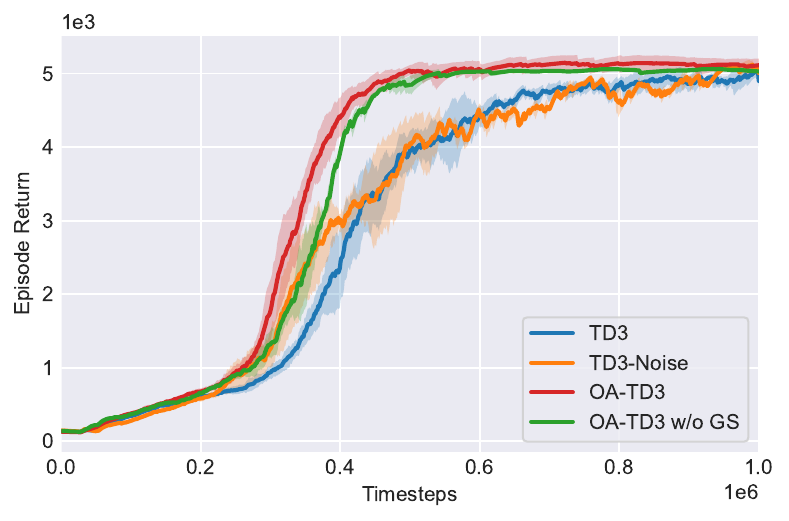}
}
\hspace{-4.1mm}
\subfigure[\emph{Ant}]{
\includegraphics[width=0.249\linewidth]{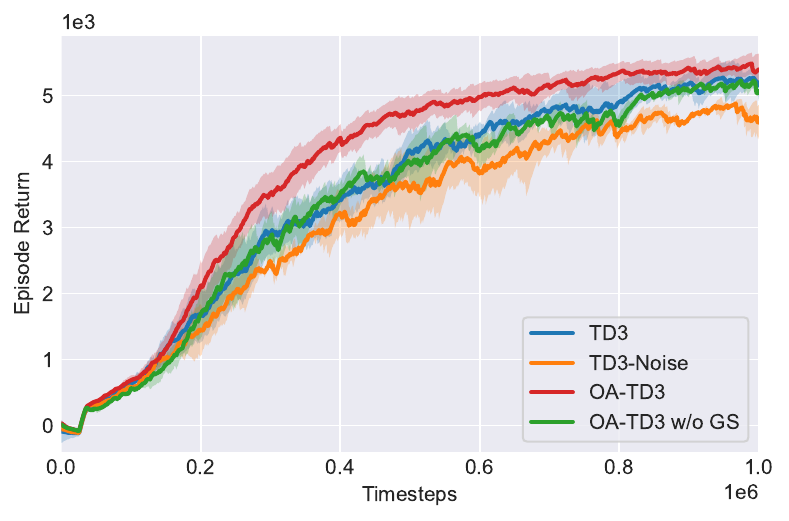}
} \\
\hspace{-4.1mm}
\subfigure[\emph{HalfCheetah}]{
\includegraphics[width=0.249\linewidth]{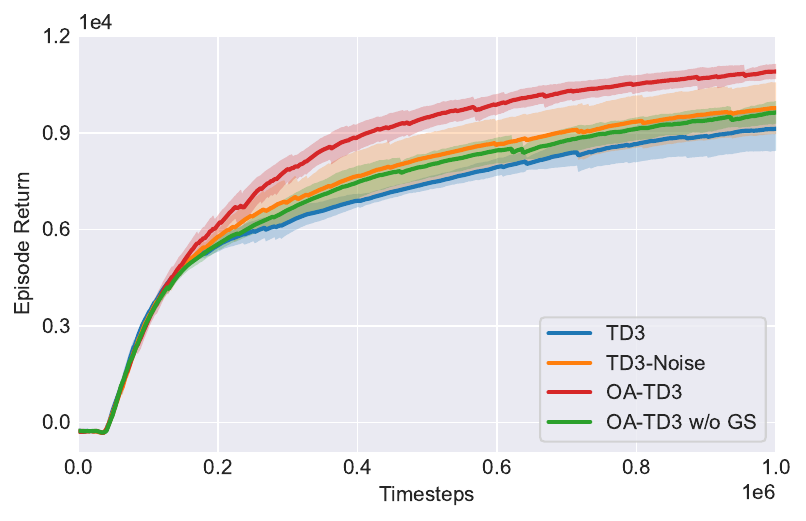}
}
\hspace{-4.1mm}
\subfigure[\emph{LunarLander}]{
\includegraphics[width=0.249\linewidth]{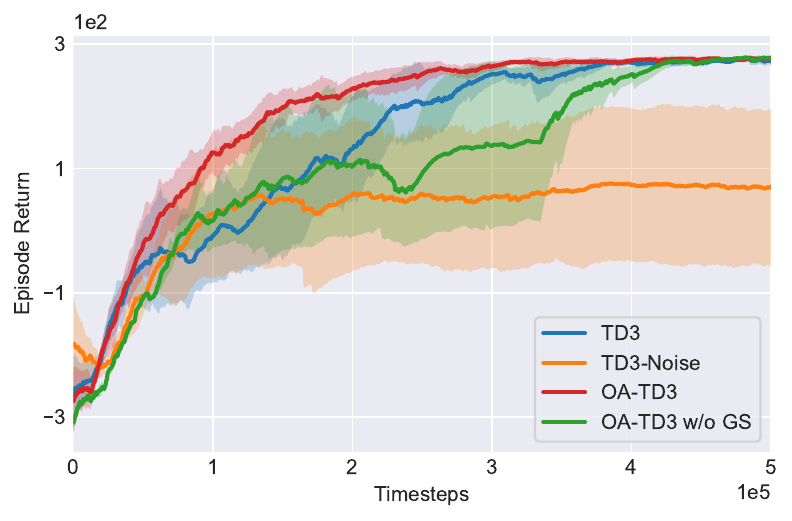}
}
\hspace{-4.1mm}
\subfigure[\emph{BipedalWalker}]{
\includegraphics[width=0.249\linewidth]{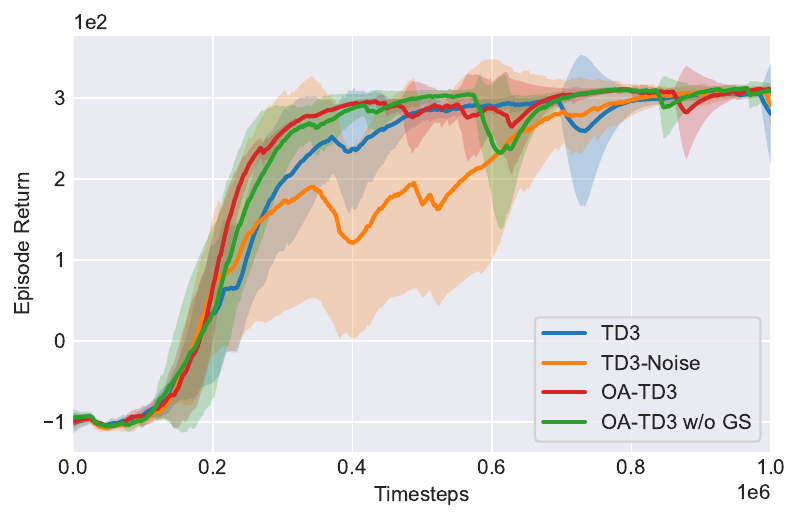}
}
\hspace{-4.1mm}
\subfigure[\emph{All tasks}]{
\includegraphics[width=0.249\linewidth]{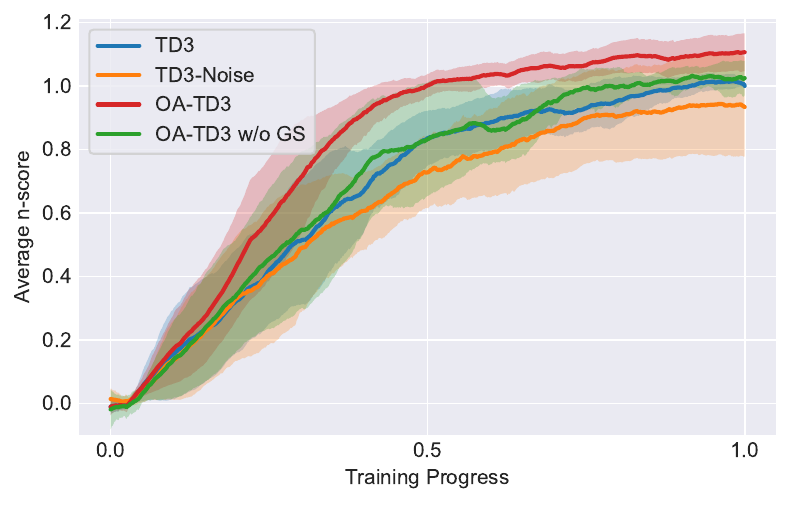}
\label{fig:all_tasks_td3}
} \\
\end{center}
\caption{
The learning curves of OA-TD3 (red) against other methods.
As depicted in the figures, OA-TD3 achieves higher sample efficiency than vanilla TD3 in most tasks.
}
\label{fig:learning_curves_td3}
\end{figure}

Besides, to analyze the influence of $\epsilon$ during evaluation, 
As results described in Table~\ref{tab:hyper_epsilon_eval_walker2d}, policies trained with $\epsilon=0.2$ are evaluated under different $\epsilon$ values in \emph{Walker2d} environment.

\begin{table}[htbp]
\caption{ Robustness of policies trained with $\epsilon=0.2$ and evaluated with different $\epsilon$ in \emph{Walker2d}.
OA-TD3 outperforms TD3 against adversaries with different perturbation strengths ($\epsilon$ values), demonstrating the effectiveness of our method.
}
\label{tab:hyper_epsilon_eval_walker2d}
\centering
\begin{tabular}{llcccc}
\toprule
$\epsilon$ (eval) & Method & Random & Biggest & Min-Q & Min-OA-Q \\ 
\midrule
\multirow{2}{*}{0.1} 
& TD3 & 3947.1 & 3835.1 & 3576.2 & 2492.1  \\ 
& OA-TD3 & \textbf{4755.3} & \textbf{4733.8} & \textbf{4709.3} & \textbf{4098.8} \\ 
\midrule
\multirow{2}{*}{0.2} 
& TD3 & 3925.4 & 3580.6 & 1938.1 & 740.2 \\ 
& OA-TD3 & \textbf{4712.7} & \textbf{4637.1} & \textbf{4005.9} & \textbf{1172.3} \\ 
\midrule
\multirow{2}{*}{0.3} 
& TD3 & 3763.5 & 2356.3 & 960.1 & 358.6 \\ 
& OA-TD3 & \textbf{4547.4} & \textbf{4389.3} & \textbf{2317.6} & \textbf{827.3} \\ 
\midrule
\multirow{2}{*}{0.4} 
& TD3 & 3465.4 & 1508.0 & 479.9 & 94.7 \\ 
& OA-TD3 & \textbf{4462.5} & \textbf{4079.3} & \textbf{940.6} & \textbf{351.1} \\ 
\bottomrule
\end{tabular}
\end{table}

\subsection{Maze2D: An Example Study}
\label{sec:app_maze2d_experiment}

\begin{figure}[htbp]
\begin{center}
\subfigure[\emph{Maze2D} environment]{
\label{fig:maze2d_env}
{\includegraphics[height=3.11cm]{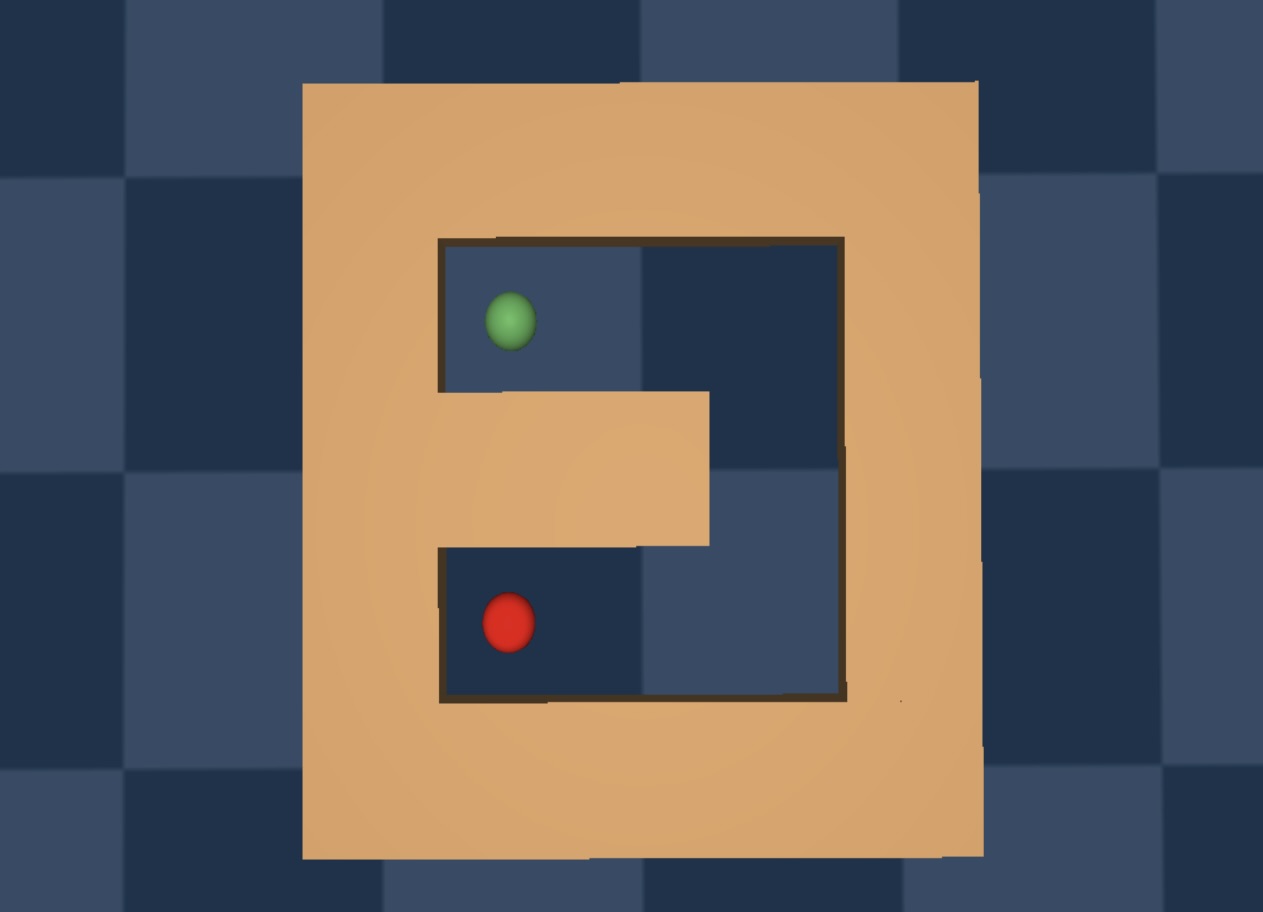}}
}
\subfigure[Result of TD3]{
\label{fig:maze2d_exp_td3}
\includegraphics[height=3.2cm]{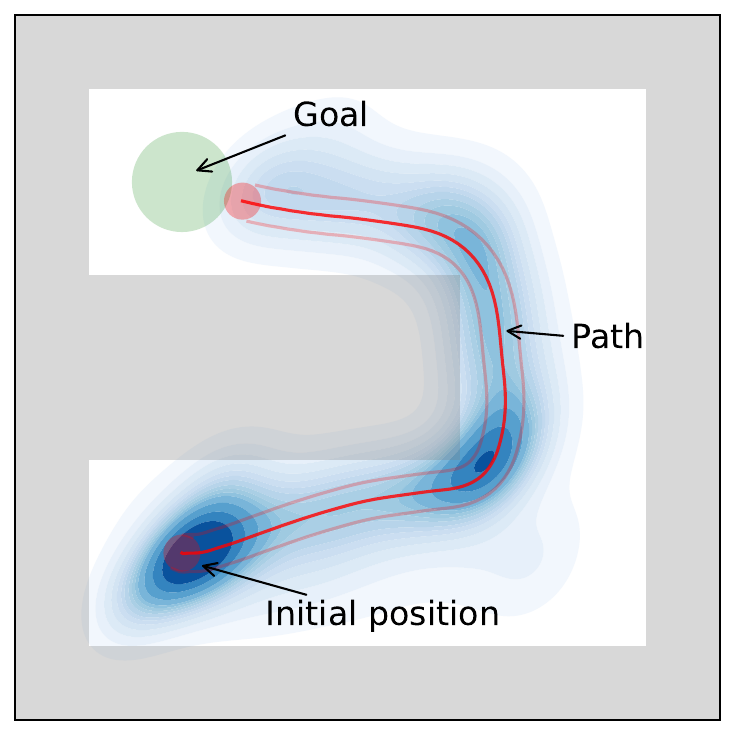}
}
\subfigure[Result of OA-TD3]{
\label{fig:maze2d_exp_oa_td3}
\includegraphics[height=3.2cm]{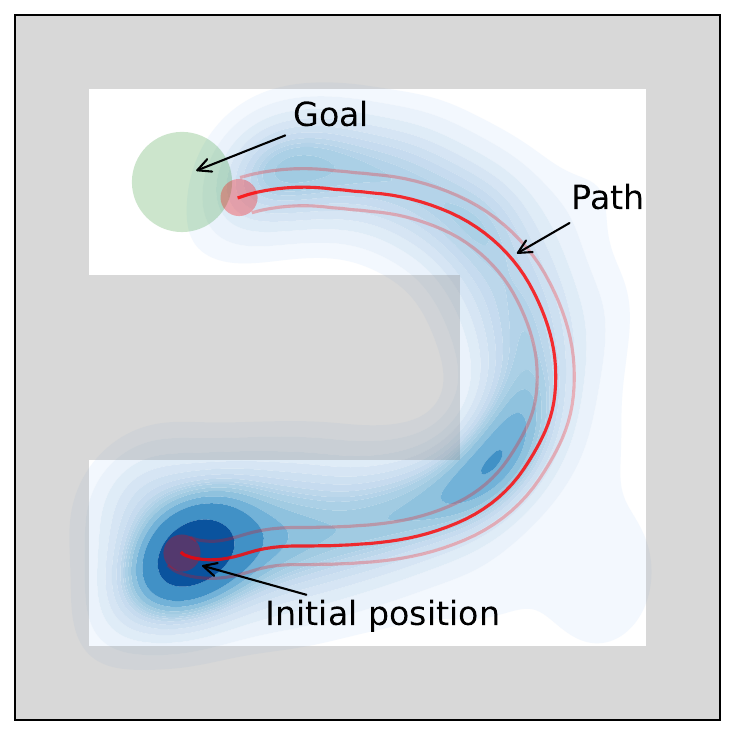}
}
\end{center}
\caption{
The experiment results on the \emph{Maze2D} environment.
(a) The visualization of \emph{Maze2D} environment.
(b) The path found by the TD3 policy.
(c) The path found by the OA-TD3 policy.
}
\label{fig:maze2d_exp_res}
\end{figure}

In this section, we analyze the difference between OA-TD3 and typical TD3 in a simple \emph{Maze2D} task.
As shown in Fig.~\ref{fig:maze2d_env}, a red ball is required to reach the target goal (green region) from the initial position (lower left corner) in a closed maze, where gray areas denote impassable walls.
The 2-DoF ball is actuated by the forces in the x and y directions.
This environment is constructed based on the \emph{Maze2D} task~\cite{fu2020d4rl} and \emph{PointMaze} environment implemented by Gymnasium Robotics.

\textbf{Observation:}
The observation $o \in \mathbb{R}^4$, where $o_{1:2}$ denote x and y coordinates of the red ball in the maze.
$o_{3:4}$ denote the linear velocity of the ball in the x and y directions.

\textbf{Action:}
The red ball is force-force-actuated and simulated based on the MuJoCo engine.
The action $a \in [-1, 1]^2$ are the linear forces exerted on the red ball in the x and y directions.
The ball velocity is clipped in a range of 5 m/s in order for it not to grow unbounded.

\textbf{Reward:}
The agent is given a reward 200 if the ball is in the goal region.
The ball is considered to have reached the goal if the Euclidean distance between both is less than 0.5 m.

The experiment results are shown in Fig.~\ref{fig:maze2d_exp_td3} and \ref{fig:maze2d_exp_oa_td3}, where the red lines represent the final path discovered by the algorithm.
The blue background displays the distribution of ball positions during exploration, which is visualized using kernel density estimation (KDE).

As shown in the figures, the results of two algorithms are quite different.
As depicted in Fig.~\ref{fig:maze2d_exp_td3}, TD3's path closely follows the wall to optimize goal achievement.
However, the ball following this trajectory may collide with walls due to action perturbations, resulting in decreased rewards or even task failure.
In contrast, the path taken by OA-TD3, as depicted in Fig.~\ref{fig:maze2d_exp_oa_td3}, maintains a safe distance from the walls, making it more resilient to potential action perturbations compared to TD3.

\end{document}